\documentclass[sigconf,nonacm]{acmart}

\usepackage{microtype}
\usepackage{graphicx}
\usepackage{subcaption}
\usepackage{booktabs} 
\usepackage{array}
\usepackage{hyperref}
\usepackage{multirow}
\usepackage{url}
\usepackage{xurl}

\usepackage{amsmath}
\usepackage{mathtools}
\usepackage{amsthm}
\newcommand{\etal}{\textit{et al.}}

\theoremstyle{plain}
\newtheorem{theorem}{Theorem}[section]
\newtheorem{proposition}[theorem]{Proposition}

\newtheorem{corollary}[theorem]{Corollary}
\theoremstyle{definition}

\theoremstyle{remark}

\usepackage[ruled,vlined]{algorithm2e}
\usepackage{pbalance}
\usepackage{marvosym}
\usepackage[utf8]{inputenc}
\usepackage[most]{tcolorbox} 
\usepackage{xcolor}
\definecolor{promptgray}{RGB}{245, 245, 245}
\definecolor{bordergray}{RGB}{180, 180, 180}

\newtcolorbox{systempromptbox}[1]{
    colback=promptgray,       
    colframe=bordergray,      
    fonttitle=\bfseries,      
    coltitle=black,           
    title=#1,                 
    arc=4pt,                  
    outer arc=4pt,
    boxrule=0.8pt,            
    left=10pt, right=10pt,    
    top=8pt, bottom=8pt,
    enhanced,                 
    attach boxed title to top left={xshift=10pt, yshift=-8pt},
    boxed title style={
        colback=white, 
        colframe=bordergray, 
        arc=2pt,
        boxrule=0.5pt
    }
}

\AtBeginDocument{%
  }

\begin{document}

\title{When Physical Preferences Meet Semantic Constraints: \\Physical and Semantic Direct Preference Optimization for Text-to-Video Generation}

\author{Siwei Meng}
\orcid{0009-0008-3044-2842}
\affiliation{%
  \institution{Computer Science and Engineering}
  \institution{University of Nevada, Reno}
  \city{Reno}
  \state{NV}
  \country{USA}
}
\email{siweim@unr.edu}

\author{Yawei Luo}
\affiliation{%
  \institution{School of Software Technology}
  \institution{Zhejiang University}
  \city{Hangzhou}
  \state{Zhejiang}
  \country{China}}
\email{yaweiluo@zju.edu.cn}

\author{Shu Zhang}
\affiliation{%
  \institution{Independent Researcher}
  \city{San Jose}
  \state{CA}
  \country{USA}
}
\email{iamshuzhang@gmail.com}

\author{Ping Liu\textsuperscript{\Letter}}
\affiliation{%
  \institution{Computer Science and Engineering}
  \institution{University of Nevada, Reno}
  \city{Reno}
  \state{NV}
  \country{USA}
}
\email{pino.pingliu@gmail.com}

\renewcommand{\shortauthors}{Meng et al.}

\begin{abstract}
Text-to-video (T2V) generation models have achieved strong visual realism, but improving physical plausibility can come at the cost of semantic consistency with the input text.
This tension arises because physical preference is typically determined by comparing dynamics between two videos, without accounting for whether either video faithfully depicts the scene specified by the prompt, making physical-semantic conflict a systematic tendency under this supervision paradigm.
We formulate this challenge as a constrained preference optimization problem and propose Physical and Semantic Direct Preference Optimization (PSDPO), which modulates each preference pair's contribution based on the agreement between its physical and semantic signals.
A gradient-level analysis shows that PSDPO bounds the semantic drift from conflicting pairs to a controllable residual, and further motivates a staged optimization protocol that provably reduces cumulative drift.
The resulting method operates entirely within the standard DPO framework, requiring no auxiliary models or additional loss terms.
Experiments show that PSDPO improves physical plausibility by up to $2\times$ over the baseline on VideoPhy-2, while maintaining strong semantic consistency on VBench, achieving a more reliable balance than existing preference-based methods.
\end{abstract}
\keywords{Video Generation, Physical Commonsense, Direct Preference Optimization, Preference Alignment}


\begin{teaserfigure}
\centering
  \includegraphics[width=1.0\textwidth]{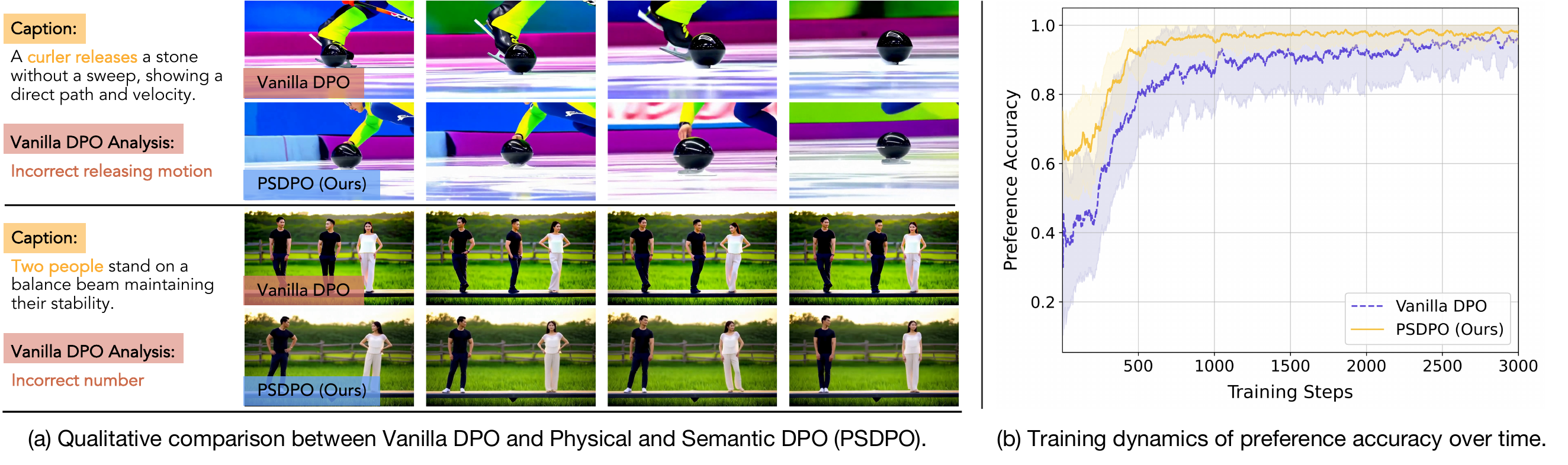}
  \vspace{-16pt}
  \captionsetup{font=small}
  \caption{
  Failure mode of preference-based physical alignment and comparison of preference accuracy.
  (a) 
  Vanilla DPO is trained using physically preferred pairs without accounting for semantic drift, which leads to videos that are physically more plausible yet semantically misaligned with the prompt.
  Our proposed PSDPO incorporates semantic feasibility into physical preference optimization, which helps preserve prompt semantics while maintaining physical plausibility.
  (b) 
  Compared to vanilla DPO, PSDPO achieves faster convergence and reduced oscillation in preference accuracy, reflecting more stable learning under heterogeneous supervision signals.
  }
  \label{fig:failuremode}
  \vspace{2pt}
\end{teaserfigure}


\maketitle

\renewcommand{\thefootnote}{}
\footnotetext{\Letter\ Corresponding author.}
\renewcommand{\thefootnote}{\arabic{footnote}}

\section{Introduction}

Text-to-video (T2V) generation~\cite{kong2024hunyuanvideo_arxiv2024,hacohen2024ltx_arxiv2024,wan2025wan_arxiv2025,yang2024cogvideox_iclr2025} has achieved impressive visual realism, yet generated videos can still violate basic physical principles~\cite{meng2025grounding_ijcai2025,kang2024far_icml2025, ding2025understanding}.
Preference-based alignment, particularly Direct Preference Optimization (DPO)~\cite{rafailov2023direct_nips2023}, has emerged as a promising approach for improving physical plausibility by learning from pairwise comparisons~\cite{qian2025rdpo,zhang2025think_arxiv2025,le2025gravity_arxiv2025, ji2025physmaster_arxiv2025,Zhang2026PhysRVGPU_arxiv2026,yuan2026inference_arxiv2026,liu2025videodpo_cvpr2025, wu2025densedpo_arxiv2025, huang2025vistadpo_icml2025, jiang2025huvidpo_arxiv2025,mai2025contextanyone}.
However, as illustrated in Figure~\ref{fig:failuremode}, when preference-based methods are directly used to optimize physical plausibility, they often cause a pronounced degradation in semantic adherence to the input text, leading to unstable and slower convergence~\cite{qian2025rdpo, hao2025enhancing_arxiv2025}.


Our analysis traces this behavior to a structural asymmetry in supervision.
Physical plausibility is more reliably assessed through relative comparisons: both humans and automated judges compare two videos and determine which one better obeys physical laws~\cite{christiano2017deep_nips2017,liu2024evalcrafter_cvpr2024}.
Absolute physical scoring, by contrast, shows limited agreement with human judgments (Table~\ref{tab:pce_human}), partly because absolute ratings tend to produce conservative, middle-range scores that obscure subtle physical differences~\cite{zheng2023judging_nips2023}.
Semantic consistency, on the other hand, is more naturally treated as a per-sample evaluation, where each video is independently checked against the input text using alignment models~\cite{radford2021learning_icml2021,hessel2021clipscore_emnlp2021,wang2024internvid_iclr2024}.
This exposes a structural asymmetry: physical judgments are typically comparative, whereas semantic constraints are per-sample feasibility requirements.

\begin{figure*}
  \includegraphics[width=1.0\textwidth]{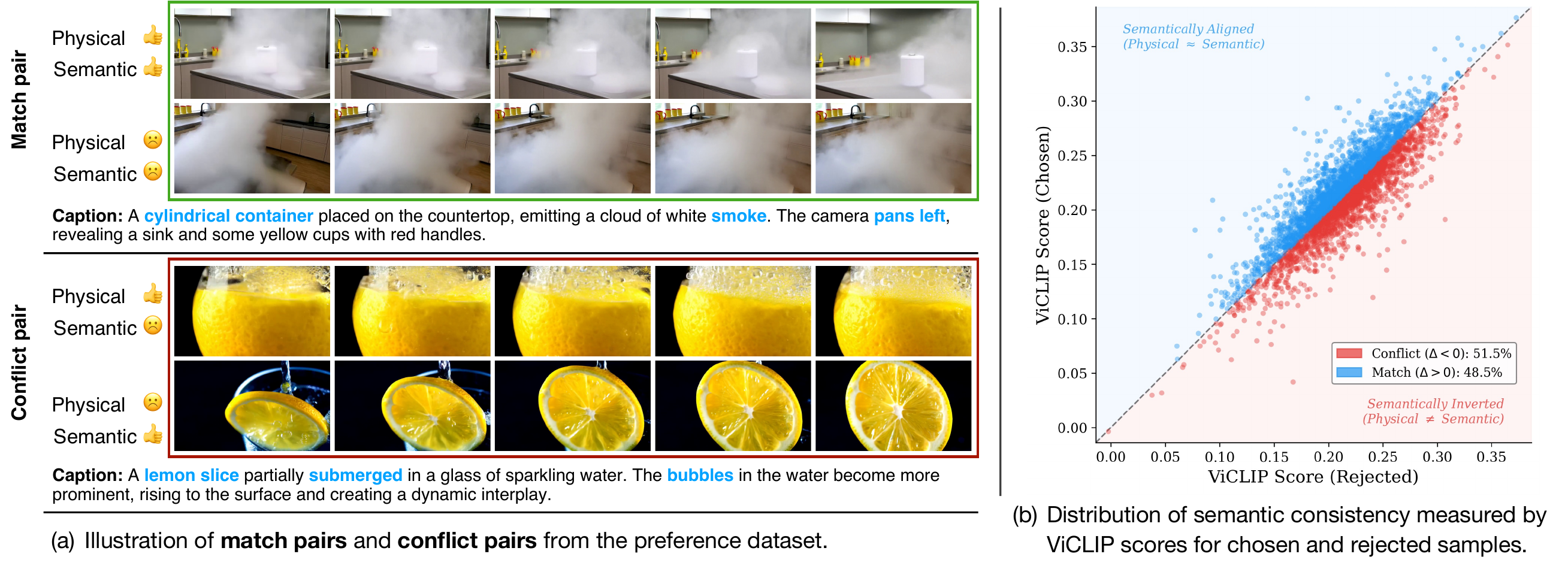}
  \vspace{-16pt}
  \captionsetup{font=small}
  \caption{Match and conflict pairs in the preference dataset and the semantic score distribution.
  (a) Preference pairs from the dataset under varying semantic drift. The \textit{lemon slice} pair shows that physically preferred videos are not always better aligned with the prompt semantics, highlighting the need for semantic-aware preference optimization.
  (b) Blue points correspond to match pairs, where the physical preferred sample also has a higher semantic score.
  The red points refer to conflict pairs with inverse preference, which occupy more than half of the preference dataset, bringing biased optimization signals during training.
  }
  \label{fig:conflictpair}
\end{figure*}

Standard preference optimization overlooks this asymmetry and implicitly assumes that physically preferred samples are also semantically valid~\cite{wang2025physcorr_arxiv2025,qian2025rdpo}.
However, because physical preference is typically determined by comparing dynamics between two videos~\cite{liu2025videodpo_cvpr2025,wang2025physcorr_arxiv2025,qian2025rdpo}, the evaluation focuses on which video better obeys physical laws and does not explicitly account for whether either video faithfully depicts the objects, actions, or scene layout specified by the prompt.
As illustrated in the left panel of Figure~\ref{fig:conflictpair}, the physically preferred ``lemon slice motion” sample better captures the intended physical motion, such as the ``submerge'' action and ``rising bubbles'', however, it depicts a whole lemon rather than a lemon slice, making it semantically less aligned than the physically rejected sample.
This mismatch occurs in over half of physical preference pairs, causing optimization to favor physical dynamics at the expense of semantic alignment and leading to semantic drift.


This diagnosis points to a natural formulation: physical preference optimization should account for whether the physically preferred video also preserves semantic consistency with the input text prompt.
Based on this formulation, we propose Physical and Semantic Direct Preference Optimization (PSDPO), which reconciles physical preferences with semantic constraints within a unified optimization objective.
PSDPO selectively adjusts how each preference pair contributes to learning based on the agreement between its physical and semantic signals.
We provide a gradient-level analysis showing that unconstrained preference optimization introduces systematic directional bias from conflicting pairs, whereas PSDPO bounds this drift to a controllable residual (Proposition~\ref{prop:bound}).
We further analyze why this drift is most severe in early training, and derive a staged optimization strategy accordingly.
Based on the formal analysis underlying PSDPO, the constrained formulation, gradient bounds, and staged protocol all operate entirely within the standard DPO framework, requiring no auxiliary models, additional loss terms, or changes to the base architecture.
This demonstrates that our principled formulation can yield both theoretical soundness and practical simplicity.
Across VBench~\cite{huang2024vbench_cvpr2024}, VideoPhy-2~\cite{bansal2025videophy_arxiv2025}, and PhyGenBench~\cite{meng2024towards_icml2025}, PSDPO consistently outperforms existing preference-based alignment methods.
For example, PSDPO achieves a VBench score of 84.13 (+1.13 improvement with the same backbone) and doubles physical commonsense performance on VideoPhy-2.

In summary, our contributions are:
\vspace{-5pt}
\begin{itemize}
    \item We identify and explain a structural failure mode in preference-based physical alignment for T2V generation: because physical preference is determined by relative comparison without accounting for semantic fidelity, physical-semantic conflict arises as a systematic tendency, not a data artifact.
    \item We formulate physical alignment as a constrained preference optimization problem and propose PSDPO, which integrates semantic feasibility into preference learning through a modulation mechanism derived from the constrained formulation.
    \item We provide a gradient-level analysis that bounds the semantic drift from conflicting pairs (Proposition 3.1), and derive a staged optimization protocol that provably reduces cumulative drift (Corollary 3.2), all within the standard DPO framework without auxiliary models or additional loss terms.
    \item PSDPO consistently outperforms existing methods on VBench, VideoPhy-2, and PhyGenBench, improving physical plausibility by up to $2\times$ while preserving semantic alignment.
\end{itemize}

\section{Related Works}
\label{sec:related_works}


\subsection{Text-to-Video Generation} 
Recent T2V models based on diffusion transformers~\cite{openai2024sora,kong2024hunyuanvideo_arxiv2024,yang2024cogvideox_iclr2025,wan2025wan_arxiv2025} have achieved strong visual realism, but primarily optimize for perceptual fidelity and semantic alignment without explicitly modeling the physical dynamics implied by the prompt.
Attempts to address this limitation include PhysGen \cite{liu2024physgen_eccv2024}, PhysGaussian~\cite{xie2024physgaussian_cvpr2024}, PhysMotion \cite{tan2024physmotion_arxiv2024}, and PhyMAGIC~\cite{meng2025phymagic_eccv2026}, which incorporate explicit simulations and object-centric physical modeling, while Xue \etal~\cite{xue2025phyt2v_cvpr2025} propose physics-informed training objectives with Chain of Thought (CoT).
While effective in controlled settings, these approaches typically require predefined physical models or task-specific assumptions, which may limit their generalization to open-ended text prompts and diverse physical situations.
Our framework instead focuses on physical plausibility as an alignment objective, motivating preference-based supervision for learning physical commonsense in T2V generation.


\subsection{Video Evaluation Metrics}
Video evaluation has traditionally focused on visual quality and semantic alignment.
Visual quality metrics, such as Fr\'{e}chet Video Distance (FVD)~\cite{unterthiner2019fvd} and Inception Score (IS)~\cite{barratt2018note_arxiv2018}, assess distributional similarity between generated and real videos but are largely insensitive to temporal dynamics and physical consistency.
Semantic alignment metrics, including CLIP-Score~\cite{wang2024internvid_iclr2024} and VideoCLIP~\cite{xu2021videoclip_emnlp2021}, capture correlations between text prompts and visual content, yet often fail to detect fine-grained dynamic anomalies.
Recognizing these limitations, recent benchmarks such as PhysGenBench~\cite{chen2025hierarchical_arxiv2025} and VideoPhy~\cite{bansal2024videophy_iclr2025,bansal2025videophy_arxiv2025} have been proposed to evaluate physical commonsense in video generation.
However, when used as supervision for preference construction, absolute physical scoring shows limited agreement with human judgments (Table~\ref{tab:pce_human}), whereas relative evaluation substantially improves reliability~\cite{zheng2023judging_nips2023}.
Motivated by this, we develop a Physical Commonsense Evaluation (PCE) system based on relative comparison with auxiliary geometric grounding, achieving 75.54\% agreement with human judgments.


\subsection{Preference Alignment in Generative Models}

Direct Preference Optimization (DPO)~\cite{rafailov2023direct_nips2023} and its variants have been widely adopted for aligning generative models with human intent, and have recently been extended to video generation for improving visual quality and temporal coherence~\cite{liu2025videodpo_cvpr2025,wu2025densedpo_arxiv2025}.
PhysCorr~\cite{wang2025physcorr_arxiv2025} and RDPO~\cite{qian2025rdpo} further apply preference alignment to physical dynamics, with PhysCorr combining physical and semantic reward signals and RDPO leveraging real-world video data for preference construction.
Both methods treat all preference pairs as equally valid supervision, implicitly assuming that the physically preferred sample is also semantically consistent with the prompt.
As discussed in Section~\ref{subsec:formulation}, this assumption is systematically violated under the standard paradigm of relative physical comparison, affecting over half of all preference pairs in practice.
PSDPO addresses this gap at the formulation level by treating semantic consistency as a feasibility condition within the preference optimization objective.

\section{Method}
\label{sec:method}
 
We address the problem of preference-based physical alignment for text-to-video generation, where the supervision signals for physical plausibility and semantic consistency are structurally different.
As empirically illustrated in Figure~\ref{fig:failuremode},
directly applying standard preference optimization to physical alignment induces semantic drift, where the generated videos progressively deviate from the textual description despite improving in physical plausibility.
In this section, we first formalize this tension as a constrained optimization problem (Section~\ref{subsec:formulation}), then derive a tractable objective that naturally resolves the conflict (Section~\ref{subsec:psdpo}), provide a gradient-level justification (Section~\ref{subsec:gradient}), and  analyze how drift accumulates over training to derive a principled optimization protocol (Section~\ref{subsec:staged}).

\begin{figure*}[ht]
    \centering
    \includegraphics[width=0.88\linewidth]{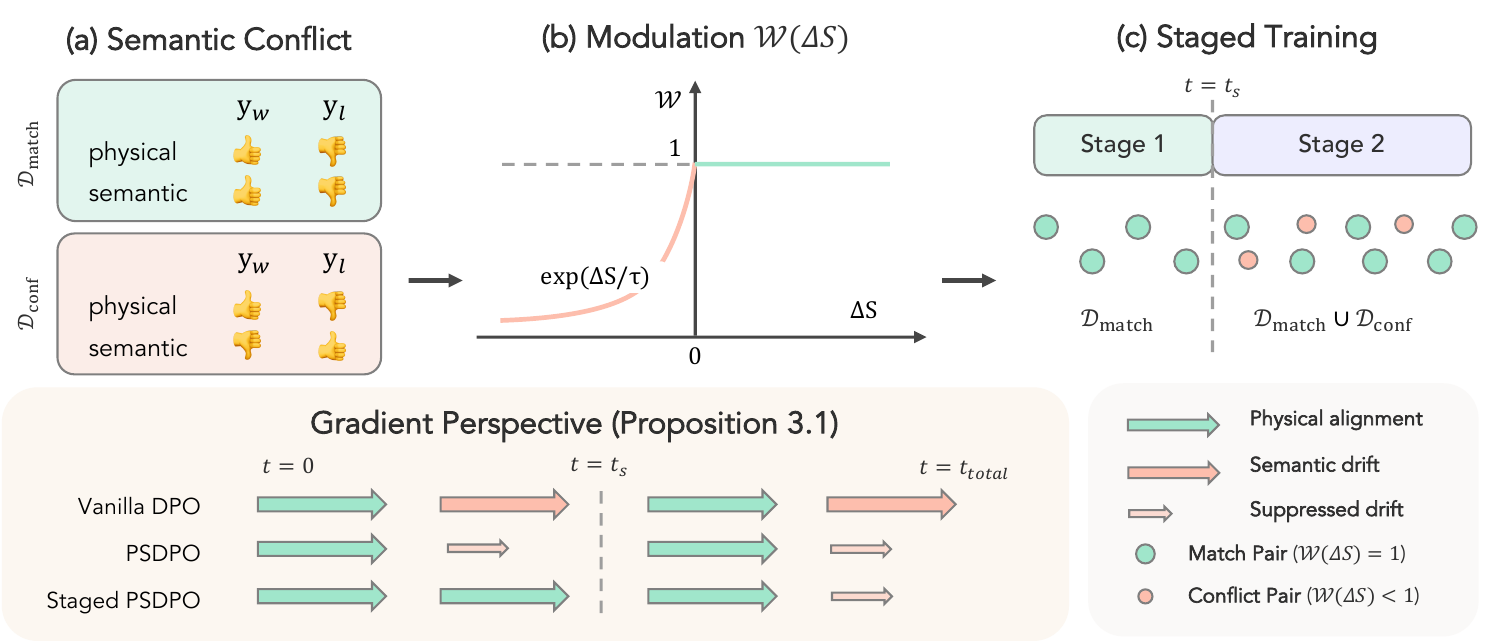}
    \vspace{-5pt}
    \captionsetup{font=small}
    \caption{
    Overview of the PSDPO framework. (a) Preference pairs are classified as semantically matching ($\Delta S\geq 0$) or conflicting ($\Delta S < 0$) based on text-video alignment scores. (b) An asymmetric modulation function $\mathcal{W}(\Delta S)$ reweights each pair's contribution to the DPO loss, preserving consistent pairs while suppressing conflicting ones proportionally. 
    (c) PSDPO Training follows a two-stage learning: the first stage optimizes on $\mathcal{D}_{\mathrm{match}}$ only, and the second stage introduces $\mathcal{D}_{\mathrm{conf}}$ with modulated weights to extract residual physical signal without semantic drift.
    }
    \Description{}
    \label{fig:psdpo_modulation}
\end{figure*}

\subsection{Problem Formulation: Constrained Physical Preference Optimization}
\label{subsec:formulation}
 
Consider a text-to-video model parameterized by $\theta$, a reference policy $\pi_{\mathrm{ref}}$ obtained via supervised fine-tuning, and a preference dataset $\mathcal{D} = \{(x, y_w, y_l)\}$ where $y_w \succ y_l$ denotes that $y_w$ is physically preferred over $y_l$ given prompt $x$.
Standard DPO optimizes:
\begin{align}
    \min_\theta \;  \mathcal{L}_{\text{DPO}} = - \mathbb{E}_{(x, y_w, y_l) \sim \mathcal{D}} \bigg[ & \log \sigma \bigg( \beta \log \frac{\pi_\theta(y_w|x)}{\pi_{\text{ref}}(y_w|x)} \nonumber \\
    & - \beta \log \frac{\pi_\theta(y_l|x)}{\pi_{\text{ref}}(y_l|x)} \bigg) \bigg].
    \label{eq:vanilla_dpo}
\end{align}
This formulation implicitly assumes that the physically preferred sample $y_w$ is also semantically valid with the prompt $x$.
However, this assumption is frequently violated in practice.
Empirical analysis reveals that approximately 51.50\% (Figure~\ref{fig:conflictpair}) of physical preference pairs exhibit semantic inversion, where the physically preferred video scores lower on text-video alignment than the rejected one.
When such pairs dominate the gradient, optimization is driven toward physically plausible but semantically inconsistent solutions.
 
To account for this, we formulate physical alignment as a constrained preference optimization problem.
Let $S_{\mathrm{sem}}(y, x)$ denote the semantic alignment score of video $y$ with respect to prompt $x$, and the objective is:
\begin{equation}
\begin{aligned}
\min_\theta \quad & \mathcal{L}_{\mathrm{DPO}}(\theta) \\
\text{s.t.} \quad & S_{\mathrm{sem}}(y_w, x) \geq S_{\mathrm{sem}}(y_l, x), \quad \forall (x, y_w, y_l) \in \mathcal{D}.
\end{aligned}
\label{eq:constrained}
\end{equation}
Concretely, we seek to optimize physical preferences while ensuring that the preferred sample also preserves semantic fidelity.
A naive approach would discard all violating pairs, but this can eliminate a substantial portion of the training data.
More importantly, the physical preference labels in these conflicting pairs remain correct: the preferred video is indeed more physically plausible, and these pairs still provide useful gradient signal for physical alignment~\cite{ren2018learning_ciml2018,han2018co_nips2018}.
We therefore seek a mechanism that preserves their physical supervision while reducing their semantic interference.

\subsection{Physical and Semantic DPO}
\label{subsec:psdpo}
 
As established in Section~\ref{subsec:formulation}, the constrained formulation in Eq.~\ref{eq:constrained} partitions preference pairs into consistent pairs ($\Delta S \geq 0$) and conflicting pairs ($\Delta S < 0$), where $\Delta S = S_{\mathrm{sem}}(y_w, x) - S_{\mathrm{sem}}(y_l, x)$ denotes the semantic discrepancy.
Hard filtering by the indicator $\mathbf{1}_{\{\Delta S \geq 0\}}$ would discard all conflicting pairs and lose their physical supervision signal.
We instead seek a smooth, strictly positive relaxation $\mathcal{W}(\Delta S)$ that preserves their contribution proportionally, satisfying the following properties:
\begin{enumerate}
    \item[(D1)] \textbf{Preserve consistent pairs:} When $\Delta S \geq 0$ (physical and semantic signals agree), $\mathcal{W}(\Delta S) = 1$, recovering standard DPO.
    \item[(D2)] \textbf{Monotonic suppression:} When $\Delta S < 0$, the weight should decrease monotonically as the semantic conflict grows.
    \item[(D3)] \textbf{Soft attenuation:} Pairs with mild conflicts ($\Delta S$ slightly below zero) should still contribute, as they carry useful physical information. Only severely conflicting pairs should be strongly suppressed.
    \item[(D4)] \textbf{Non-negativity:} $\mathcal{W}(\Delta S) > 0$ for all $\Delta S$, so no pair is entirely discarded and every pair contributes a nonzero training signal.
\end{enumerate} 

Properties D1--D4 restrict the admissible family of weighting functions $\mathcal{W}$.
D1 requires $\mathcal{W}(\Delta S) = 1$ on the consistent region $\Delta S \geq 0$; D2 requires monotonic decay for $\Delta S < 0$; D4 requires strict positivity, which rules out linear forms such as $\max(0, 1 + \Delta S / \tau)$ that truncate to zero; and D3 favors smooth attenuation near the boundary.
A simple function satisfying all four properties is the asymmetric exponential:
\begin{equation}
    \mathcal{W}(\Delta S) =
    \begin{cases} 
    1.0 & \text{if } \Delta S \ge 0, \\
    \exp\!\left(\frac{\Delta S}{\tau}\right) & \text{if } \Delta S < 0,
    \end{cases}
    \label{eq:reweighting}
\end{equation}
where $\tau > 0$ is a temperature controlling the attenuation rate.
We empirically compared this with the linear decay variant and found that the exponential form yields sharper DPO margins and faster convergence (see the training-curve comparison in the supplementary material).

The temperature $\tau$ governs a continuous interpolation between two extremes.
As $\tau \to 0^+$, $\mathcal{W}(\Delta S) \to 0$ for all $\Delta S < 0$, effectively discarding all conflicting pairs and recovering the hard constraint in Eq.~\ref{eq:constrained}.
As $\tau \to \infty$, $\mathcal{W}(\Delta S) \to 1$ for all $\Delta S$, and PSDPO reduces to vanilla DPO.
Intermediate values of $\tau$ realize a smooth trade-off: retaining useful physical signal from mildly conflicting pairs while suppressing severely conflicting ones.
This interpolation also provides robustness to noise in the semantic scorer $S_{\mathrm{sem}}$: a moderately large $\tau$ prevents the modulation from overreacting to small scoring errors, while still suppressing pairs with clear semantic violations.
Section~\ref{subsec:gradient} provides a formal characterization of how $\tau$ controls the residual drift.

Integrating $\mathcal{W}(\Delta S)$ into the DPO objective yields the PSDPO loss:
\begin{align}
    \mathcal{L}_{\text{PSDPO}} = - \mathbb{E}_{(x, y_w, y_l) \sim \mathcal{D}} \bigg[ & \mathcal{W}(\Delta S) \cdot \log \sigma \bigg( \beta \log \frac{\pi_\theta(y_w|x)}{\pi_{\text{ref}}(y_w|x)} \nonumber \\
    & - \beta \log \frac{\pi_\theta(y_l|x)}{\pi_{\text{ref}}(y_l|x)} \bigg) \bigg],
    \label{eq:psdpo_loss}
\end{align}
where $\pi_\theta$ denotes the trainable policy, $\pi_{\text{ref}}$ is the frozen SFT-initialized reference model, and $\beta$ controls the strength of preference optimization.
 
The resulting objective preserves a direct correspondence with the constrained formulation in Eq.~\ref{eq:constrained}.
For pairs satisfying the constraint ($\Delta S \geq 0$), $\mathcal{W} = 1$ and the loss is identical to standard DPO.
For pairs violating the constraint ($\Delta S < 0$), $\mathcal{W} < 1$ and the loss contribution is automatically reduced in proportion to the severity of the violation.
This transforms the constrained problem into a single weighted objective without auxiliary models, additional loss terms, or changes to the DPO optimization framework.
Although the resulting mechanism takes the form of per-sample reweighting, it differs from generic importance weighting used in robust optimization or domain adaptation: the weight is not derived from distributional shift or label noise, but from the structural conflict between two heterogeneous supervision signals within each preference pair, and its form is guided by the desiderata D1--D4.

\subsection{Gradient Analysis under Semantic Conflicts}
\label{subsec:gradient}
 
We provide a gradient-level analysis to explain why standard DPO drifts and why PSDPO corrects this behavior.
We partition the dataset as $\mathcal{D} = \mathcal{D}_{\mathrm{match}} \cup \mathcal{D}_{\mathrm{conf}}$, where $\mathcal{D}_{\mathrm{match}}$ contains pairs with $\Delta S \geq 0$ and $\mathcal{D}_{\mathrm{conf}}$ contains pairs with $\Delta S < 0$.
Under standard DPO, the expected gradient decomposes as:
\begin{equation}
\mathbb{E}[\nabla_{\theta}\mathcal{L}_{\mathrm{DPO}}]
=
\underbrace{\mathbb{E}_{\mathcal{D}_{\mathrm{match}}}[\nabla_\theta \ell]}_{\text{Desired alignment}}
+
\underbrace{\mathbb{E}_{\mathcal{D}_{\mathrm{conf}}}[\nabla_\theta \ell]}_{\text{Drift bias}},
\label{eq:grad_decomp}
\end{equation}
where $\ell$ denotes the per-sample DPO loss.
The second term systematically pushes the model toward physically preferred but semantically invalid solutions, producing semantic drift.
Under PSDPO, each pair in $\mathcal{D}_{\mathrm{conf}}$ is scaled by $\mathcal{W}(\Delta S) = \exp(\Delta S / \tau)$, yielding:
\begin{equation}
\mathbb{E}[\nabla_{\theta}\mathcal{L}_{\mathrm{PSDPO}}]
=
\mathbb{E}_{\mathcal{D}_{\mathrm{match}}}[\nabla_\theta \ell]
+
\epsilon,
\label{eq:psdpo_grad}
\end{equation}
where the residual $\epsilon$ is bounded as follows.
 
\begin{proposition}[Residual gradient bound]
\label{prop:bound}
Let $G = \sup \| \nabla_\theta \ell \|$ denote the uniform gradient bound.
Then:
\begin{equation}
\| \epsilon \| \;\leq\; G \cdot \mathbb{E}_{\mathcal{D}_{\mathrm{conf}}} \!\left[ \exp\!\left(\frac{\Delta S}{\tau}\right) \right].
\label{eq:eps_bound}
\end{equation}
Since $\Delta S < 0$ on $\mathcal{D}_{\mathrm{conf}}$, the exponential factor is strictly less than 1, and decays rapidly as $\tau \to 0^+$.
In the limit $\tau \to 0^+$, $\| \epsilon \| \to 0$ and PSDPO recovers optimization over $\mathcal{D}_{\mathrm{match}}$ alone.
\end{proposition}
 
The bound confirms that the temperature $\tau$ directly controls how much drift bias leaks into the gradient.
In practice, $\tau = 0.02$ reduces the drift contribution to a negligible residual while retaining useful gradient signal from mildly conflicting pairs (Table~\ref{tab:ablationTemp}).

\subsection{Controlling Drift Accumulation during Training}
\label{subsec:staged}
 
Although PSDPO modulates preference gradients via~$\mathcal{W}(\Delta S)$, Proposition~\ref{prop:bound} bounds only the per-step residual $\epsilon$ from $\mathcal{D}_{\mathrm{conf}}$.
Over the course of training, these per-step residuals accumulate and can collectively steer the model away from semantic consistency.
A direct consequence of Proposition~\ref{prop:bound} is that a staged training strategy formally reduces this cumulative drift.
 
Let $B = G \cdot \mathbb{E}_{\mathcal{D}_{\mathrm{conf}}}[\mathcal{W}(\Delta S)]$ denote the per-step residual bound from Proposition~\ref{prop:bound}, which is a constant independent of $\theta$ since $G = \sup_\theta \|\nabla_\theta \ell\|$ is a uniform bound and $\mathcal{W}(\Delta S)$ depends only on the data.
We compare two training protocols: direct optimization, which trains on the full dataset $\mathcal{D} = \mathcal{D}_{\mathrm{match}} \cup \mathcal{D}_{\mathrm{conf}}$ for all $T_{\mathrm{total}}$ steps, and staged optimization, which trains on $\mathcal{D}_{\mathrm{match}}$ alone for the first $T_{\mathrm{s}}$ steps before incorporating $\mathcal{D}_{\mathrm{conf}}$.
 
\begin{corollary}[Cumulative residual reduction]
\label{cor:staged}
Let $E^{\mathrm{direct}} = \sum_{t=1}^{T_{\mathrm{total}}} \epsilon^{(t)}$ and $E^{\mathrm{staged}} = \sum_{t=T_{\mathrm{s}}+1}^{T_{\mathrm{total}}} \epsilon^{(t)}$ denote the cumulative residual under direct and staged training, respectively, where $\epsilon^{(t)}$ is the residual gradient from $\mathcal{D}_{\mathrm{conf}}$ at step $t$.
By the triangle inequality and Proposition~\ref{prop:bound}:
\begin{equation}
\| E^{\mathrm{staged}} \| \leq (T_{\mathrm{total}} - T_{\mathrm{s}}) \cdot B, \quad \| E^{\mathrm{direct}} \| \leq T_{\mathrm{total}} \cdot B.
\label{eq:staged_bound}
\end{equation}
Staged training reduces the cumulative residual bound by a factor of $T_{\mathrm{s}} / T_{\mathrm{total}}$.
\end{corollary}
 

Unlike standard curriculum learning \cite{bengio2009curriculum_icml2009,9392296_pami2022}, which orders samples by difficulty to accelerate convergence, the staging here is determined by the constrained formulation: $\mathcal{D}_{\mathrm{match}}$ and $\mathcal{D}_{\mathrm{conf}}$ correspond to feasible and infeasible pairs in Eq.~\ref{eq:constrained}, and Corollary~\ref{cor:staged} shows that excluding $\mathcal{D}_{\mathrm{conf}}$ in early training provably reduces the cumulative semantic drift bound.
Based on this analysis, we adopt a two-stage training protocol.
At training step $t$, the PSDPO objective is optimized over:
\begin{equation}
\begin{aligned}
\label{eq:staged_psdpo}
    \mathcal{L}_{\mathrm{PSDPO}}^{(t)} = - & \mathbb{E}_{(x,y_w,y_l)\sim \mathcal{D}^{(t)}} \Bigg[ 
      \mathcal{W}(\Delta S) \cdot \log \sigma \\
      & \bigg( \beta \log \frac{\pi_\theta(y_w|x)}{\pi_{\mathrm{ref}}(y_w|x)} 
      - \beta \log \frac{\pi_\theta(y_l|x)}{\pi_{\mathrm{ref}}(y_l|x)} 
    \bigg) \Bigg],
\end{aligned}
\end{equation}
where the training set follows a step-function curriculum with switching threshold $T_{\mathrm{s}}$:
\begin{equation}
\mathcal{D}^{(t)} =
\begin{cases}
    \mathcal{D}_{\mathrm{match}}, & \text{if } t < T_{\text{s}}, \\
    \mathcal{D}_{\mathrm{match}} \;\cup\;  \mathcal{D}_{\mathrm{conf}}, & \text{otherwise}.
\end{cases}
\label{eq:Dt_curriculum}
\end{equation}
As outlined in Algorithm~\ref{alg:psdpo_curr}, the first stage ($t < T_{\mathrm{s}}$) builds a reliable preference representation on unambiguous pairs with zero drift accumulation.
The second stage introduces conflicting pairs with their contributions modulated by $\mathcal{W}(\Delta S)$, enabling the model to extract residual physical signal from semantically ambiguous data in a controlled manner.
Table~\ref{tab:ablationComponent} confirms that staged training improves semantic fidelity from 77.08 to 78.25.

\begin{algorithm}[t]
\caption{Staged PSDPO Training.}
\label{alg:psdpo_curr}
\KwIn{
Preference dataset $\mathcal{D} = \{(x, y_w, y_l, \Delta S)\}$; \\
Initial policy $\pi_\theta$; reference policy $\pi_{\text{ref}}$; learning rate $\eta$; \\
DPO temperature $\beta$; semantic temperature $\tau$; \\
Total training steps $T_{\text{total}}$; switching step $T_{\text{s}}$.
}
\KwOut{Optimized policy $\pi_\theta$}
Define semantic-aware weight: 
Eq.~\ref{eq:reweighting}

Partition dataset into two subsets: $\mathcal{D}_{\mathrm{match}}$ and $\mathcal{D}_{\mathrm{conf}}$

\For{$t = 1$ \KwTo $T_{\text{total}}$}{
    
    \eIf{$t < T_{\text{s}}$}{
        Construct batch by sampling from $\mathcal{D}_{\mathrm{match}}$\;
    }{

        Construct batch by sampling from
        $\mathcal{D}_{\mathrm{match}} \;\cup\; \mathcal{D}_{\mathrm{conf}}$\;
    }
    
    Compute PSDPO loss $\mathcal{L}_{\mathrm{PSDPO}}^{(t)}$ with Eq.~\ref{eq:staged_psdpo}\;
    
    Update $\theta \leftarrow \theta - \eta \nabla_\theta \mathcal{L}_{\mathrm{PSDPO}}^{(t)}$\;
}

\end{algorithm}

\section{Experiments}
\label{sec:exp}

\begin{figure*}[t]
    \centering
    \includegraphics[width=0.98\linewidth]{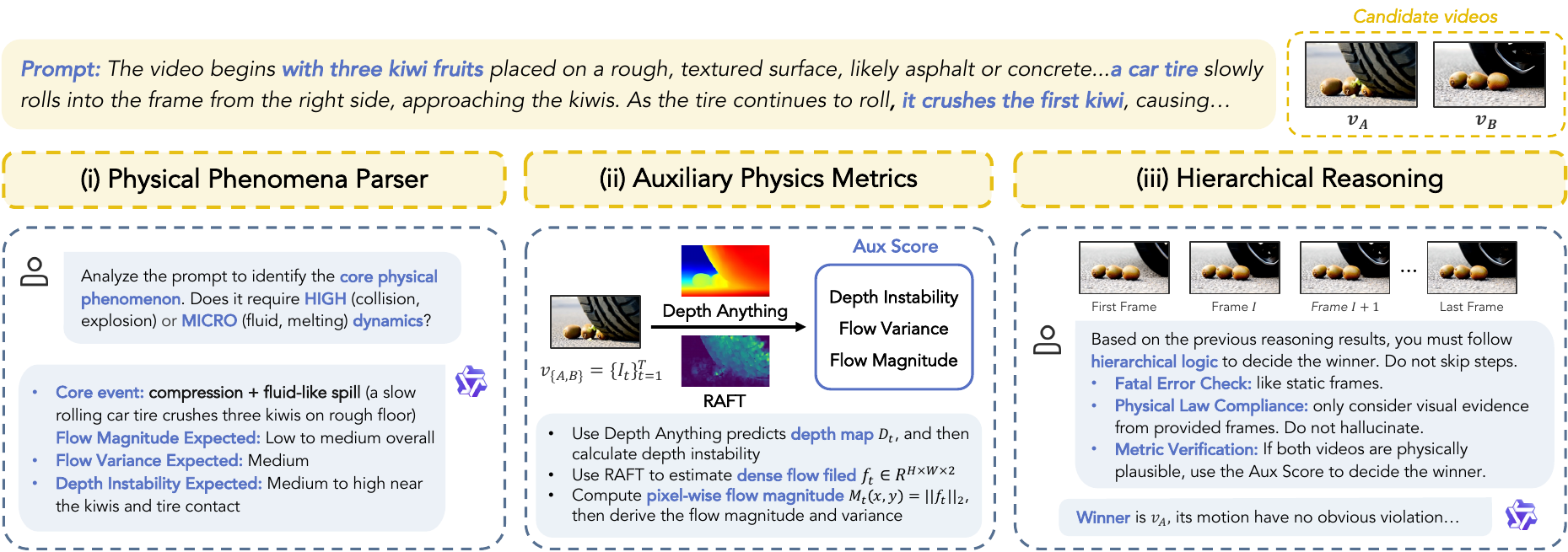}
    \vspace{-8pt}
    \captionsetup{font=small}
    \caption{\textbf{The architecture of Physical Commonsense Evaluation.}
    PCE evaluates physical plausibility through a hierarchical pipeline that combines (i) physical phenomena parsing, (ii) auxiliary physics metrics, and (iii) hierarchical reasoning.
    }
    \label{fig:physicalEvaluation}
    \Description{}
\end{figure*}

\subsection{Preference Dataset Construction}
\label{subsec:dataset_construction}

\paragraph{Dataset Construction}
For supervised fine-tuning, we sample 10K text-video pairs from WISA-80K~\cite{wang2025wisa_arxiv2025}, covering 17 common physical events.
These pairs are used for supervised fine-tuning of Wan-1.3B model, producing the SFT-initialized model that serves as both the starting point of PSDPO training and the frozen reference policy.
We then reuse the 10K sampled captions to construct the preference dataset.
For each caption, we generate two candidate videos from the SFT-initialized model by varying the random seed and diffusion noise, yielding 20K candidate videos.
Each pair is annotated with a relative physical preference by our Physical Commonsense Evaluation (PCE) system, while each individual video is assigned an absolute semantic feasibility score $S_\mathrm{sem}$ by calculating the text-video similarity with ViCLIP~\cite{wang2024internvid_iclr2024}.
These two signals together form the final 10K preference pairs used for PSDPO training.

\paragraph{Physical Commonsense Evaluation (PCE) System}
Off-the-shelf MLLMs~\cite{yang2025qwen3} struggle to manage information flow when facing mismatched text-video pairs, often leading to biased judgments where visual artifacts corrupt semantic reasoning. 
To mitigate this modality interference, we propose the Physical Commonsense Evaluation (PCE) system. 
Our PCE system is built upon the open-source Qwen3-VL model~\cite{yang2025qwen3} as the underlying MLLM judge. 
Unlike relying on holistic scoring, PCE employs a hierarchical reasoning pipeline that disentangles semantic expectation from visual execution. 
This framework, as illustrated in Figure~\ref{fig:physicalEvaluation}, comprises three core components: (i) the \textbf{Physical Phenomena Parser}, (ii) \textbf{Auxiliary Geometric Metrics}, and (iii) MLLM-based \textbf{Hierarchical Reasoning} module. 
Crucially, this design enforces that physical ground truths are established solely from text before interacting with potentially noisy visual data.
A detailed description of the PCE system is provided in the supplementary material.


\subsection{Implementation details}
We implement PSDPO on the Wan-1.3B architecture~\cite{wan2025wan_arxiv2025} and fine-tune the model with Low-Rank Adaptation (LoRA)~\cite{hu2022lora_iclr2022} to preserve pre-trained priors.
LoRA is applied to the query, key, value, and output projections of the attention blocks as well as the feed-forward layers.
Following empirical observations on video generation stability, we set the DPO temperature parameter to $\beta=2000$.
The total training step $T_{\text{total}}$ is 3k, and the switching step $T_{\text{s}}$ is 1k.
Training is conducted on 4 NVIDIA RTX 6000 Pro GPUs using the AdamW optimizer~\cite{loshchilov2017decoupled_arxiv2017} with a learning rate of $1\mathrm{e}{-5}$, while the reference policy $\pi_{\mathrm{ref}}$ is fixed as the SFT-initialized Wan-1.3 model.

\begin{table}[t]
\centering
\footnotesize
\setlength{\abovecaptionskip}{1pt}
\captionsetup{font=small}
\caption{Agreement with human judgments on physical plausibility under different evaluation paradigms.
}
\label{tab:pce_human}
\resizebox{\linewidth}{!}{
\begin{tabular}{l c c c}
\toprule
\textbf{Evaluator} &
\textbf{Evaluation Type} &
\textbf{Supervision} &
\textbf{Agreement (\%)} \\
\midrule

MLLM (Qwen3-VL) &
Absolute &
Visual reasoning &
15.06  \\

MLLM (Qwen3-VL) &
Relative &
Visual reasoning &
66.48  \\

PCE &
Absolute &
Visual reasoning + geometry cues &
18.06 \\

\midrule
PCE (Ours) &
Relative &
Visual reasoning + geometry cues &
\textbf{75.54} \\
\bottomrule
\end{tabular}}
\end{table}

\begin{table}[t]
\centering
\footnotesize
\setlength{\abovecaptionskip}{1pt}
\captionsetup{font=small}
\caption{
\textbf{Quantitative Comparison Results on VBench.}
We compare PSDPO with representative preference-based and supervised alignment methods.
PSDPO consistently improves overall video quality while preserving semantic fidelity.
PSDPO w/o staged denotes PSDPO training with single stage. 
}
\label{tab:vbench}
\resizebox{\linewidth}{!}{
\begin{tabular}{l c c c c}
\toprule
\textbf{Methods} & \textbf{Base Models} & \textbf{Total}$(\uparrow)$ & \textbf{Quality}$(\uparrow)$ & \textbf{Semantics}$(\uparrow)$ \\
\midrule
VideoDPO & CogVideoX-2B & 79.80 & 83.00 & 66.99 \\
PhysHPO & CogVideoX-2B & 82.50 & 83.30 & \textbf{79.30} \\
\midrule
Omni-Video & Wan-1.3B & 83.00 & 84.27 & 77.92 \\
PSDPO w/o staged & Wan-1.3B & \underline{83.69} & \underline{85.35} & 77.08 \\
\textbf{PSDPO} & Wan-1.3B & \textbf{84.13} & \textbf{85.60} & \underline{78.25} \\
\bottomrule
\end{tabular}
}
\end{table}

\begin{table}[t]
\centering
\footnotesize
\setlength{\abovecaptionskip}{1pt}
\captionsetup{font=small}
\caption{
\textbf{Quantitative Comparison Results on VideoPhy-2 Benchmark.}
We compare base text-to-video models and alignment methods under Hard, Activity, and Interaction subsets.
Staged PSDPO achieves the best performance on Hard and Activity physical subset and yields competitive overall results.
}
\label{tab:videophy2}
\resizebox{\linewidth}{!}{
\begin{tabular}{l c c c c c}
\toprule
\textbf{Methods} & \textbf{Base Models} & \textbf{Hard}$(\uparrow)$ & \textbf{Activity}$(\uparrow)$ & \textbf{Interaction}$(\uparrow)$ & \textbf{Overall}$(\uparrow)$ \\
\midrule
Wan-1.3B & -- & 0.039 & 0.241 & 0.220 & 0.234 \\
\midrule
ProPhy & Wan-1.3B & 0.072 & -- & -- & \textbf{0.265} \\
PSDPO w/o staged & Wan-1.3B & 0.056 & 0.243 & \textbf{0.270} & 0.250 \\
\textbf{PSDPO} & Wan-1.3B & \textbf{0.078} & \textbf{0.257} & \underline{0.258} & \underline{0.258}  \\
\bottomrule
\end{tabular}
}
\end{table}

\begin{table}[t]
\centering
\footnotesize
\setlength{\abovecaptionskip}{1pt}
\captionsetup{font=small}
\caption{
\textbf{Quantitative Comparison Results on PhyGenBench Benchmark.}
We report category-wise physical consistency across Mechanics, Optics, Thermal, and Material dimensions.
PSDPO achieves strong and balanced improvements across all categories, matching or surpassing specialized physics-aware methods.
}
\label{tab:phygenbench}
\resizebox{\linewidth}{!}{
\begin{tabular}{lc c c c c}
\toprule
\textbf{Methods} & \textbf{Mechanics}$(\uparrow)$ & \textbf{Optics}$(\uparrow)$ & \textbf{Thermal}$(\uparrow)$ & \textbf{Material}$(\uparrow)$ & \textbf{Average}$(\uparrow)$ \\
\midrule
CogVideoX-2B & 0.38 & 0.43 & 0.34 & 0.39 & 0.39 \\
+ PhysHPO& 0.50 & 0.56 & 0.47 & \textbf{0.58} & 0.53 \\
\midrule
CogVideoX-5B & 0.43 & 0.55 & 0.40 & 0.42 & 0.45 \\
+ DiffPhy& \textbf{0.53} & 0.59 & \textbf{0.58} & 0.46 & \textbf{0.54} \\
\midrule
Wan-1.3B& 0.44 & 0.60 & 0.41 & 0.42 & 0.47 \\
+ \textbf{PSDPO}& \textbf{0.53} & \textbf{0.62} & \underline{0.52} & \underline{0.47} & \textbf{0.54} \\
\bottomrule
\end{tabular}
}
\end{table}

\begin{table}[t]
\centering
\footnotesize
\setlength{\abovecaptionskip}{1pt}
\captionsetup{font=small}
\caption{\textbf{Quantitative results on VBench and VideoPhy.}
We compare multi-stages of our alignment pipeline on VBench and VideoPhy-2 (5-point).
Vanilla DPO improves physical realism but degrades semantic fidelity.
PSDPO w/o staged improves semantic alignment with semantic-aware reweighting, while staged PSDPO further stabilizes optimization and has the best overall results.
}
\label{tab:ablationComponent}
\setlength\tabcolsep{8pt}
\resizebox{\linewidth}{!}{%
\begin{tabular}{l|ccc|cc}
\toprule
\multirow{2}{*}{\textbf{Method}} 
& \multicolumn{3}{c|}{\textbf{VBench}} 
& \multicolumn{2}{c}{\textbf{VideoPhy-2}} \\
& \textbf{Total}$(\uparrow)$ & \textbf{Quality}$(\uparrow)$ & \textbf{Semantics}$(\uparrow)$ 
& \textbf{SA}$(\uparrow)$ & \textbf{PC}$(\uparrow)$ \\
\midrule
Baseline        & 83.31 & 85.23 & 76.65 & 3.073 & 3.675 \\
SFT             & 83.10 & 84.25 & \underline{77.51} & 3.063 & 3.743 \\
Vanilla DPO     & 82.79 & 84.23 & 77.01 & 3.073 & 3.755 \\
PSDPO  w/o staged          & \underline{83.69} & \underline{85.35} & 77.08 & \underline{3.083} & \underline{3.766} \\
\textbf{PSDPO}
                & \textbf{84.13} & \textbf{85.60} & \textbf{78.25} 
                & \textbf{3.130} & \textbf{3.775} \\
\bottomrule
\end{tabular}%
}
\end{table}

\subsection{Quantitative Comparison}
\label{subsec:quantitative_main}
We report quantitative comparisons with representative video generation and preference-based alignment methods.
Table~\ref{tab:vbench} summarizes performance on VBench~\cite{huang2024vbench_cvpr2024} using two base models of comparable parameter scale, CogVideoX-2B~\cite{yang2024cogvideox_iclr2025} and Wan-1.3B~\cite{wan2025wan_arxiv2025}.
Among the compared methods and backbones, PSDPO with single stage already achieves the highest total score and physical quality score among all compared methods.
With staged learning enabled, PSDPO further improves overall performance while maintaining competitive semantic alignment.
These results indicate that progressively introducing semantically conflicting preferences stabilizes optimization and improves physical plausibility without degrading text–video consistency.

Table~\ref{tab:videophy2} reports results on VideoPhy-2~\cite{bansal2025videophy_arxiv2025} using Wan-1.3B as the base model.
Compared to the base model, all physical preference alignment methods achieve consistent improvements across evaluation subsets.
PSDPO demonstrates strong physical reasoning performance across different subsets of VideoPhy-2.
In particular, PSDPO without staged learning achieves the highest score on the Interaction subset (0.270), indicating superior modeling of multi-object interaction dynamics.
With staged learning, PSDPO further improves performance on the challenging Hard subset (0.078), yielding a $2\times$ improvement over the base model while maintaining competitive overall physical commonsense performance.



To further analyze physical commonsense across different physical events, Table~\ref{tab:phygenbench} compares PSDPO with physics-aware baselines on PhyGenBench along Mechanics, Optics, Thermal, and Material dimensions.
PSDPO achieves competitive or superior performance across all four categories.
Compared with PhysHPO, which is built on a comparable-scale base model (CogVideoX-2B), PSDPO outperforms it by a clear margin on Mechanics (+0.03) and Optics (+0.06), achieving a higher overall average (0.54 vs.\ 0.53).
Furthermore, PSDPO matches DiffPhy on the overall average (0.54) and Mechanics (0.53), while outperforming it on Optics and Material, despite DiffPhy being built on the larger CogVideoX-5B backbone.
Unlike DiffPhy, PSDPO eliminates the reliance on external physical simulators or chain-of-thought reasoning during inference, demonstrating that physical adherence can be achieved by preference optimization alone.

\subsection{Qualitative Comparison}
\label{subsec:qualitative_main}
\begin{figure*}[t]
    \centering
    \includegraphics[width=0.99\linewidth]{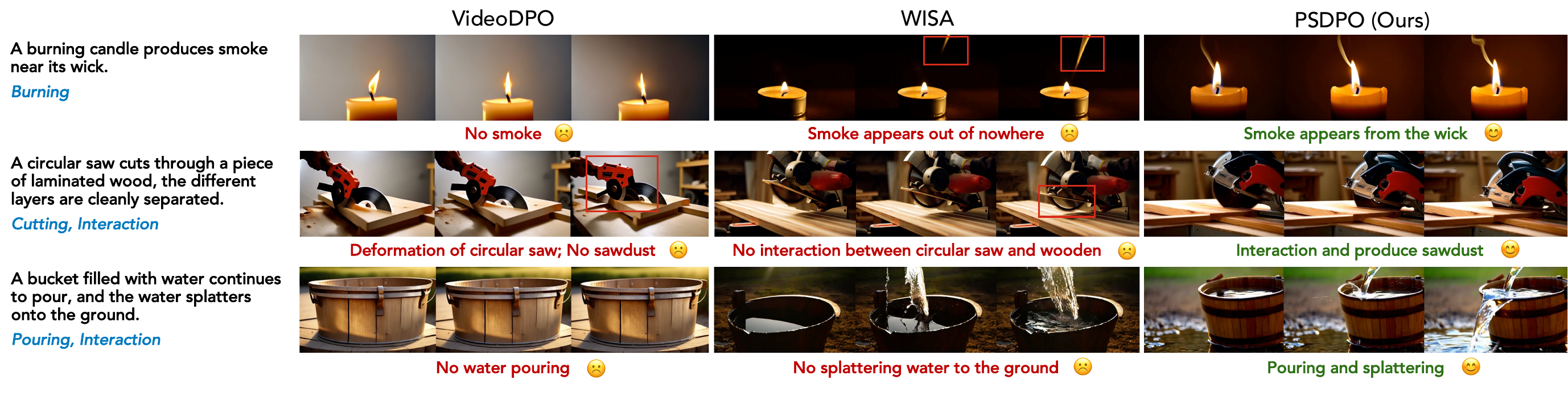}
    \setlength{\abovecaptionskip}{-0.1cm}
    \captionsetup{font=small}
    \caption{\textbf{Qualitative Comparison with alignment-based and physical-aware models.} 
    Existing methods often suffer from incorrect physical interaction or missing objects.
    In contrast, PSDPO generates physically plausible fine-gained dynamics, ensuring the objects are consistent with the text prompts.
    }
    \Description{}
    \label{fig:qualitative1}
\end{figure*}

Figure~\ref{fig:qualitative1} presents qualitative comparison with alignment-based and physical-aware methods.
VideoDPO~\cite{liu2025videodpo_cvpr2025} suffers from severe physical hallucinations, such as smoke disappearing in \textit{burning} case and object deformation in \textit{cutting wooden} case and poor semantic alignment in \textit{water pouring} case.
WISA~\cite{wang2025wisa_arxiv2025} fails to generate physically plausible scenes, such as object non-contact and disappearing. 
In contrast, PSDPO demonstrates precise physical understanding ability, such as smoke originating accurately from the wick, the sawdust producing upon contact, and water splattering realistically upon the ground.
Overall, these comparison results highlight PSDPO's superior capability to ground fine-grained physical dynamics in textual prompts, ensuring that complex interactions are generated with both physical realism and high semantic fidelity.

\begin{figure}[t]
    \centering
    \includegraphics[width=1.0\linewidth]{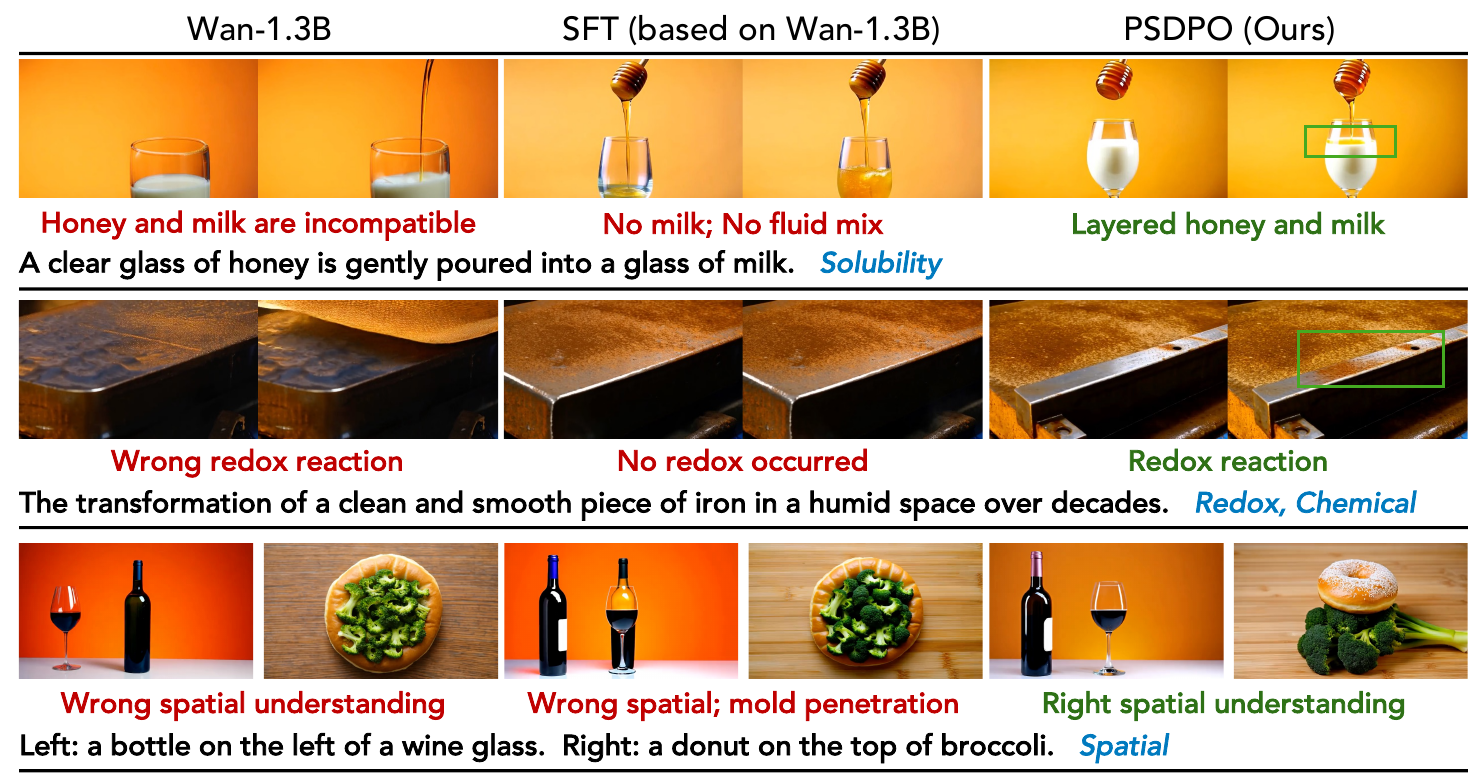}
    \setlength{\abovecaptionskip}{-0.24cm}
    \captionsetup{font=small}
    \caption{\textbf{Qualitative Comparison with baseline and SFT models.}
    Baseline and SFT models fail to capture the immiscibility of fluids and misinterpret spatial instructions, while PSDPO successfully models these physical laws and generates precise spatial constraints.
    }
    \Description{}
    \label{fig:qualitative2}
\end{figure}

Figure~\ref{fig:qualitative2} compares PSDPO with the base model (Wan-1.3B) and the SFT model, highlighting its effectiveness in correcting common physical failure modes.
In the \textit{honey and milk mix} case, both the base model and SFT fail to model fluid dynamics, producing incorrect mixing behavior, while PSDPO correctly renders the layered viscosity of fluids that obeys the immiscibility law.
In the \textit{iron redox} scenario, the baselines display a wrong redox reaction, and the SFT model has no obvious dynamic transformation.
PSDPO instead generates the gradual textural transformation characteristic, accurately capturing the rust formation process.
Finally, the \textit{spatial} example illustrates that PSDPO correctly positions objects according to the prompt, whereas baselines suffer from wrong spatial understanding or object penetration.
These cases demonstrate that PSDPO preserves object structure and spatial arrangement while maintaining physical plausibility, indicating that improvements in physical modeling do not come at the expense of semantic integrity.

\subsection{Ablation Study}
\label{subsec:ablation}
\paragraph{Impact of different components.}
Table~\ref{tab:ablationComponent} reports an ablation study on VBench and VideoPhy-2, isolating the effect of each alignment component.
Starting from the base model, we progressively add SFT, vanilla DPO, PSDPO w/o staged, and staged PSDPO.
SFT improves semantic alignment on VBench but yields limited gains in physical commonsense and slightly degrades perceptual quality, indicating that supervised signals alone are insufficient.
Vanilla DPO enhances physical realism but causes a clear drop in semantic fidelity, revealing semantic drift when physical preferences are optimized without constraints.
In contrast, PSDPO without staged training recovers semantic performance while maintaining strong physical alignment.
Finally, full PSDPO achieves the best overall results across both benchmarks, improving all VBench metrics and attaining the highest SA and PC scores on VideoPhy-2.
The results show that semantic-aware preference modulation, together with staged optimization scheduling, stabilize learning under heterogeneous supervision and leads to a more reliable balance between physical plausibility and semantic fidelity.

\paragraph{Impact of different semantic asymmetric modulation scales.}
Table~\ref{tab:ablationTemp} studies the effect of semantic modulation scale, which control the suppressing strength for PSDPO training.
Without semantic modulation, vanilla DPO prioritizes physical preference, achieving strong PC score and Quality while suffering from semantic degradation on both VBench and VideoPhy-2.
After introducing semantic modulation, PSDPO achieves improved Semantic and SA scores than vanilla DPO, while preserving physical realism.
Overall, $\tau=0.02$ achieves the best balance between semantic adherence and physical learning, and is used as the default setting in staged PSDPO training.

\begin{table}[t]
\centering
\footnotesize
\setlength{\abovecaptionskip}{1pt}
\captionsetup{font=small}
\caption{\textbf{Ablation on Semantic Asymmetric Modulation Scale.} 
We observe that setting $\lambda_{sem}, \tau$$=$$0.02$ provides an optimal balance, significantly recovering semantic adherence on VBench while maintaining physical plausibility on VideoPhy-2 (5-point).
$\lambda_{sem}$ means introducing semantic asymmetric modulation into PSDPO.
}
\label{tab:ablationTemp}
\resizebox{\linewidth}{!}{%
\begin{tabular}{l|ccc|cc}
\toprule
\multirow{2}{*}{\textbf{Method}} & \multicolumn{3}{c|}{\textbf{VBench}} & \multicolumn{2}{c}{\textbf{VideoPhy-2}} \\
 & \textbf{Total}$(\uparrow)$ & \textbf{Quality}$(\uparrow)$ & \textbf{Semantics}$(\uparrow)$ & \textbf{SA}$(\uparrow)$ & \textbf{PC}$(\uparrow)$ \\ \midrule
Vanilla DPO (w/o $\lambda_{sem}$) & 82.79 & 84.23 & 77.01 & 3.073 & \underline{3.755} \\
PSDPO (w/ $\lambda_{sem}, \tau=0.10$) & 82.94 & 85.90 & 73.61 & 3.091 & 3.738 \\
PSDPO (w/ $\lambda_{sem}, \tau=0.05$) & \textbf{83.94} & \textbf{85.87} & \underline{76.22} & \textbf{3.112} & 3.709 \\
PSDPO (w/ $\lambda_{sem}, \tau=0.02$) & \underline{83.69} & \underline{85.35} & \textbf{77.08} & \underline{3.083} & \textbf{3.766} \\
\bottomrule
\end{tabular}%
}
\end{table}

\begin{figure}
    \centering
    \includegraphics[width=1.0\linewidth]{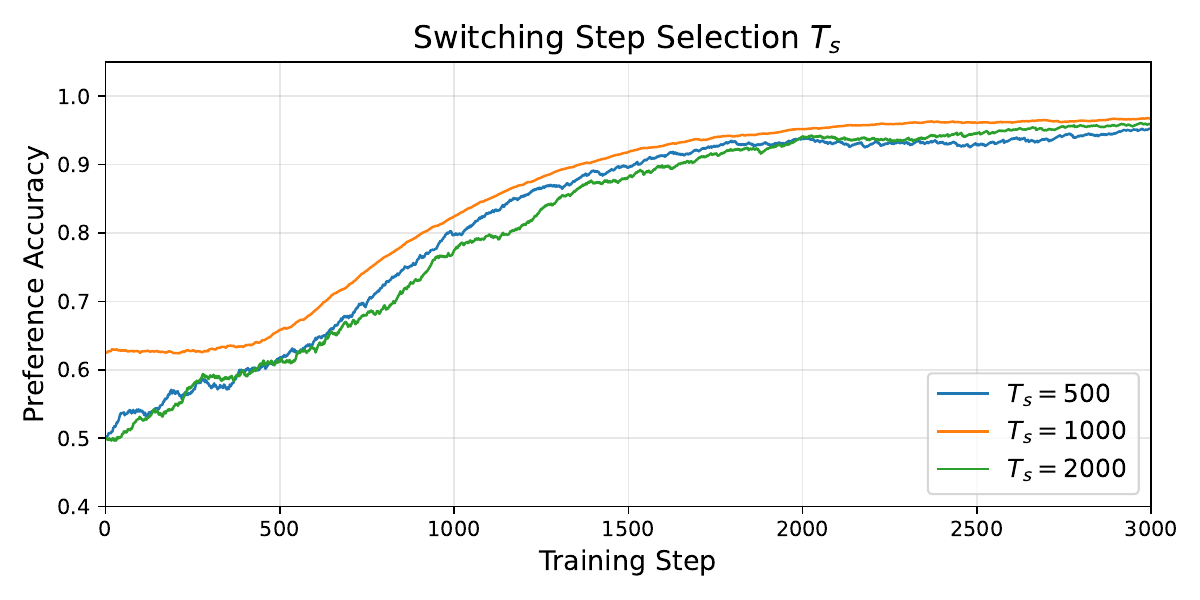}
    \vspace{-22pt}
    \caption{Preference accuracy during training under different switching steps $T_s$. $T_s=1000$ balances early stability on $\mathcal{D}_{\mathrm{match}}$ with sufficient second-stage learning from $\mathcal{D}_{\mathrm{conf}}$.}
    \label{fig:ablationTs}
\end{figure}

\paragraph{Impact of switching time step selection.}
We further study the effect of the switching step $T_{\mathrm{s}}$ in staged PSDPO training.
We fix the total training step to $T_{\mathrm{total}}=3$k, under which the model is empirically observed to converge, and vary the switching point among $T_{\mathrm{s}} \in \{500,1000,2000\}$.
Figure~\ref{fig:ablationTs} examines the trade-off implied by Corollary~\ref{cor:staged}. A larger $T_{\mathrm{s}}$ reduces the upper bound of cumulative residual drift by delaying exposure to $\mathcal{D}_{\mathrm{conf}}$, but it also shortens the second stage, limiting the model’s ability to exploit residual physical signal from semantically conflicting pairs. 
When $T_{\mathrm{s}}=500$, conflicting pairs are introduced too early, before the model has learned a stable preference representation on $\mathcal{D}_{\mathrm{match}}$, resulting in insufficient control of semantic drift. 
In contrast, when $T_{\mathrm{s}}=2000$, the first stage becomes overly conservative.
Although the accumulated drift is further reduced, the shortened second stage limits gains in physical commonsense. 
Among the three strategies, $T_{\mathrm{s}}=1000$ achieves the best balance across VBench and VideoPhy-2, consistent with our analysis that staged training should suppress early drift without overly restricting later learning from residual physical supervision.


\subsection{Efficacy of the PCE System}
\label{subsec:pce_effectiveness}
Finally, we validate the reliability of our automated Physical Commonsense Evaluation system, which serves as the reward model for physical preference. 
We conducted a human study on random selected 1k video pairs ($10\%$ of the training dataset), and compared human judgments with different evaluation paradigms in Table~\ref{tab:pce_human}. 
The results reveal a clear trend that relative evaluation consistently outperforms absolute scoring for both standalone MLLMs and the PCE system.
For example, Qwen3-VL achieves only 15.06\% agreement under absolute scoring, but improves to 66.48\% under relative comparison.
Compared with standalone Qwen3-VL, our PCE system further improves this to 75.54\% by incorporating auxiliary geometric metrics. 
These improvements demonstrate that combining relative evaluation with auxiliary geometry metrics reasoning effectively reduces hallucinations and ensures robust physical assessment.


\section{Conclusion}
\label{sec:conclusion}

In this work, we study preference-based physical alignment for T2V generation and address the challenge of improving physical plausibility without inducing semantic drift under heterogeneous supervision.
We introduce a unified approach that combines robust physical preference construction with constrained preference optimization, including the Physical Commonsense Evaluation (PCE) system for reliable relative physical judgments grounded in geometric cues, and Physical and Semantic Direct Preference Optimization (PSDPO), which incorporates semantic feasibility into preference learning by suppressing gradient contributions from semantically conflicting pairs. 
Experiments on VBench, VideoPhy-2, and PhyGenBench demonstrate that PSDPO improves physical consistency while preserving strong semantic alignment and visual quality, consistently outperforming existing preference-based methods.

\section{Appendix}

\appendix

\section{Physical Commonsense Evaluation (PCE) System}
\label{appen:pce}

Off-the-shelf MLLMs struggle to manage information flow when facing mismatched text-video pairs, often leading to biased judgments where visual artifacts corrupt semantic reasoning. 
To mitigate this modality interference, we propose the Physical Commonsense Evaluation (PCE) system. 
Unlike relying on holistic scoring, PCE employs a hierarchical reasoning pipeline that disentangles semantic expectation from visual execution. 
This framework, as illustrated in Figure~\ref{fig:physicalEvaluation}, comprises three core components: (i) the \textbf{Physical Phenomena Parser}, (ii) \textbf{Auxiliary Geometric Metrics}, and (iii) MLLM-based \textbf{Hierarchical Reasoning} module. 
Crucially, this design enforces that physical ground truths are established solely from text before interacting with potentially noisy visual data.

\paragraph{\textbf{(i) Physical Phenomena Parser}}
\label{sec:parser}
The primary objective of the Physical Phenomena Parser is to abstract expected physical laws solely from the text prompt, establishing a semantic ground truth independent of visual content. 
Specifically, we employ Qwen3-VL~\cite{yang2025qwen3} to analyze linguistic cues and extract distinct physical motion events.
To facilitate quantitative verification, these events are mapped into a four-quadrant space defined by \textbf{Depth Stability} and \textbf{Flow Magnitude}. 
We provide the common physical events classification considered in our experiment, as shown in Table~\ref{tab:taxonomy}.
For instance, destructive scenes like \textit{crushing} are classified as ``High Depth Instability, High Flow Magnitude," whereas state changes like \textit{melting} typically exhibit ``Low Depth Instability, High Flow Magnitude." 
This structured categorization ensures that the subsequent evaluation is strictly anchored to the intended physics of the prompt, providing a precise reference for geometric analysis.

\begin{systempromptbox}{System Prompt: PCE System}
\small 
You are an expert Video Physics Commonsense Reasoning Judge. Your task is to evaluate two AI-generated videos (Video A and Video B) based on a Text Prompt, Auxiliary Physics Metrics and determine which one adheres better to real-world physics.

\vspace{0.5em}
\textbf{1. Physical Phenomena Parser} \\
Analyze the Text Prompt to identify the core physical phenomenon. High flow magnitude is positive for high dynamics (explosion, crash), while low motion energy is positive for low dynamics (sleeping, reflection).

\vspace{0.5em}
\textbf{2. Metric Interpretation} \\
Compare computed Auxiliary Physics Metrics relatively (Flow Magnitude, Flow Variance, Depth Instability, Motion Energy). If Motion Energy $< 0.005$, the video is effectively frozen.

\vspace{0.5em}
\textbf{3. Visual Reasoning} \\
Reason about video frames considering: Physical law consistency (Priority 1), Prompt consistency (Priority 2), and Metric alignment (Priority 3).

\vspace{0.5em}
\textbf{4. Decision Making Logic}
\begin{itemize}
    \item \textbf{Priority 1 (Static Filter):} Reject if motion is expected but video is static.
    \item \textbf{Priority 2 (Physics Filter):} Visual evidence of physics violation (e.g., gravity, ghosting) trumps metrics.
    \item \textbf{Priority 3 (Metric Tie-Breaker):} Use metrics to decide between two plausible videos.
\end{itemize}

\vspace{0.5em}
\textbf{5. Output Format} \\
Must output ONLY valid JSON including: \texttt{physical\_phenomena\_tag}, \texttt{metric\_analysis}, \texttt{visual\_defect\_A/B}, \texttt{overall\_reasoning}, and \texttt{confidence\_score}.
\end{systempromptbox}

\paragraph{\textbf{(ii) Auxiliary Geometric Metrics}}
\label{sec:metrics}

While standard MLLM possess strong semantic reasoning capabilities, they inherently struggle to quantify fine-grained dynamic properties due to the \textit{central tendency bias}~\cite{zheng2023judging_nips2023}.
To bridge this gap, we employ state-of-the-art vision models to extract objective geometric priors, serving as reliable quantitative anchors for our Physical Phenomena Parser.

\paragraph{Optical Flow as Motion Kinematics.} 
We first utilize the RAFT model~\cite{teed2020raft_eccv2020} to quantify global speed and motion turbulence with high precision.
Let $V = \{I_t\}_{t=1}^T$ denote a video sequence with $T$ frames.
For each consecutive pair $(I_t, I_{t+1})$, RAFT estimates a dense flow field $\mathbf{f}_t \in \mathbb{R}^{H \times W \times 2}$, where $\mathbf{f}_t(x,y) = (u, v)$ represents the pixel displacement.
We compute the pixel-wise flow magnitude $M_t(x,y) = \|\mathbf{f}_t(x,y)\|_2$, and then derive the \textbf{Flow Magnitude} ($\mathcal{M}_{\text{flow}}$) and \textbf{Flow Variance} ($\mathcal{V}_{\text{flow}}$):
\begin{align}
    \mathcal{M}_{\text{flow}} &= \frac{1}{T-1} \sum_{t=1}^{T-1} \left( \frac{1}{HW} \sum_{x,y} M_t(x,y) \right), \\
    \mathcal{V}_{\text{flow}} &= \frac{1}{T-1} \sum_{t=1}^{T-1} \text{Var}_{\text{spatial}}(M_t),
\end{align}
where $\mathcal{M}_{\text{flow}}$ serves as a proxy for kinetic energy, while $\mathcal{V}_{\text{flow}}$ captures the chaos or non-uniformity of the motion (e.g., determining if a ``crushing" event exhibits the expected turbulence).

\paragraph{Depth as 3D Geometry Information.}
To assess the geometric stability, we employ the monocular estimator \textbf{Depth Anything}~\cite{yang2024depth_cvpr2024}.
Given the scale ambiguity inherent in monocular depth, we perform min-max normalization on the predicted depth map $D_t$ to obtain $\hat{D}_t \in [0,1]$.
We define \textbf{Depth Instability} ($\mathcal{I}_{\text{depth}}$) as the temporal flickering of the structural geometry:
\begin{equation}
    \mathcal{I}_{\text{depth}} = \frac{1}{T-1} \sum_{t=1}^{T-1} \frac{1}{HW} \| \hat{D}_{t+1} - \hat{D}_{t} \|_1.
\end{equation}
A lower $\mathcal{I}_{\text{depth}}$ indicates that the generated content maintains consistent 3D structure over time, adhering to rigid body physics where applicable.
With these auxiliary geometric metrics, we establish an objective physical grounding that complements the semantic evaluations of MLLMs.


\paragraph{\textbf{(iii) Hierarchical Reasoning}}
\label{sec:hierarchical_reasoning}
Combining the semantic expectations $\mathcal{P}_{tag}$ and the quantitative metrics $\mathcal{M}$, we employ a MLLM-based judge to evaluate the video pair $(v_A, v_B)$ following a strict three-level priority hierarchy:

\textbf{Fatal Error Check.} 
The system first check for fundamental generation failures. 
If a prompt implies motion but the calculated Flow Magnitude falls below a negligible threshold ($\mathcal{M}_{\text{flow}}< 0.005$ ), the video is classified as a ``Static Image" and automatically rejected. This step efficiently filters trivial cases without incurring MLLM inference costs.

\textbf{Physical Law Compliance.} 
For determining validity, the MLLM performs visual reasoning to detect violations of the identified $\mathcal{P}_{tag}$. 
It looks for severe physical hallucinations, such as ``object ghosting", ``temporal inversion", or ``floating objects." 
Qualitative visual evidence of such violations takes precedence over scalar metrics.
    
\textbf{Metric-Guided Refinement.} 
In cases where both candidates are visually plausible, the decision relies on the fine-grained alignment with auxiliary geometric metrics. 
Specifically, the system selects the candidate that best matches the magnitude profile defined by $\mathcal{P}_{tag}$ (e.g., preferring higher $\mathcal{V}_{\text{flow}}$ for a ``shattering" prompt).

\paragraph{Confidence Calibration.}
Finally, the system outputs a preference label $y \in \{v_A, v_B\}$ along with a confidence score $\mathcal{C} \in [0, 1]$ to quantify the certainty of the judgment, where $\mathcal{C}$ is set to \textbf{High} (1.0) for clear physical violations or static failures, \textbf{Medium} (0.5) for metric-based tie-breakers, and \textbf{Low} (0.1) for ambiguous cases. 
During dataset construction, we apply a confidence filter to retain only high-quality preference pairs (i.e., $\mathcal{C} \ge 0.5$), ensuring the robustness of the reward signal.

\begin{table*}[t]
\centering
\caption{\textbf{Taxonomy of Physical Motion Events.} We employ an LLM to parse text prompts into four distinct quadrants based on expected geometric behaviors. This classification serves as the semantic ground truth for our hierarchical evaluation pipeline.}
\label{tab:taxonomy}
\setlength{\tabcolsep}{4mm}
\resizebox{1.0\linewidth}{!}{%
\begin{tabular}{
    >{\centering\arraybackslash}m{2.5cm} 
    >{\centering\arraybackslash}m{2.45cm} 
    m{8.4cm} 
    m{5.8cm}@{}
}
\toprule
\textbf{Depth Instability} &  \textbf{Flow Magnitude} & \textbf{Physical Definition} & \textbf{Event Examples}  \\ \midrule

Low & Low & Scenes with minimal motion or rigid objects moving slowly without structural changes. & \textit{Sleeping, Standing, Reading, Meditating} \\ \midrule

Low & High & Objects exhibiting significant motion or shape changes while maintaining surface continuity. & \textit{Melting, Pouring, Swimming, Gliding, Flowing Water, Snake Slithering} \\ \midrule

High & Low & Objects undergoing subtle but complex structural changes in place, often involving texture or internal geometry shifts. & \textit{Breathing, Pulsing, Flickering Light, Boiling (Simmering), Heartbeat} \\ \midrule

High & High & High-energy events involving rapid structural violation, occlusion, or disintegration of objects. & \textit{Crushing, Exploding, Shattering, Colliding, Breaking, Fighting} \\ \bottomrule
\end{tabular}%
}
\end{table*}

\section{Evaluation Benchmarks}
\label{appen:benchmarks}
To comprehensively assess the performance of our model in terms of physical plausibility and semantic adherence, we employ three complementary benchmarks: VBench~\cite{huang2024vbench_cvpr2024}, PhyGenBench~\cite{meng2024towards_icml2025}, and VideoPhy-2~\cite{bansal2025videophy_arxiv2025}.

\textbf{VBench} is a comprehensive evaluation suite that assesses video generation models across 16 hierarchical dimensions, decomposed into Video Quality, Semantic, and Total score. 
Unlike single-metric evaluations, VBench utilizes a suite of specialized vision models to calculate dimension-specific scores. 
For instance, it employs RAFT~\cite{teed2020raft_eccv2020} to estimate the degree of dynamics in synthesized videos and MUSIQ~\cite{ke2021musiq_iccv2021} for image quality predictor.

\textbf{PhyGenBench} is specifically designed to evaluate physical hallucinations across four core categories: Mechanics, Optics, Thermal, and Material properties. 
The evaluation typically relies on rigorous VLM-based judgments to verify whether the generated visual dynamics comply with the physical principles implied by the text prompts.

\textbf{VideoPhy-2} is a challenging action-centric benchmark focusing on complex human-object interactions and dynamic events. 
It evaluates models on three subsets: Hard, Activity, and Interaction. 
The core metric is the Joint Performance score, which considers a video valid only if it simultaneously satisfies Semantic Adherence (SA) and Physical Commonsense (PC).
VideoPhy-2 calculates the final SA and PC scores using a 5-point scale, selecting from the following options: {Very Unlikely (1), Unlikely (2), Neutral (3), Likely (4), Very Likely (5)}.
In our experiment, we compute a joint preference which measures the fraction of videos that both adhere to the text prompt ($SA \ge 4$) and follow physical commonsense to a high degree ($PC \ge 4$), and report the joint score ($P C \ge 4|SA \ge 4$) in Table 3.
We report the average SA and PC score in Table 5, ensuring a robust assessment of dynamic physical understanding. 

\section{Semantic Modulation Form Selection}
\label{appen:semanticForm}
We conducted a comparative ablation study to justify our choice of the exponential modulation function $\mathcal{W}(\Delta S) = \exp(\frac{\Delta S}{\tau})$ for handling semantic conflicts against a linear decay variant.
The linear variant is defined as $\mathcal{W}_{linear}(\Delta S) = \max(0, 1 + \frac{\Delta S}{\tau})$, which provides a softer penalty for semantically conflicting preference pairs.

\begin{figure}[t]
    \centering
    \includegraphics[width=1.0\linewidth]{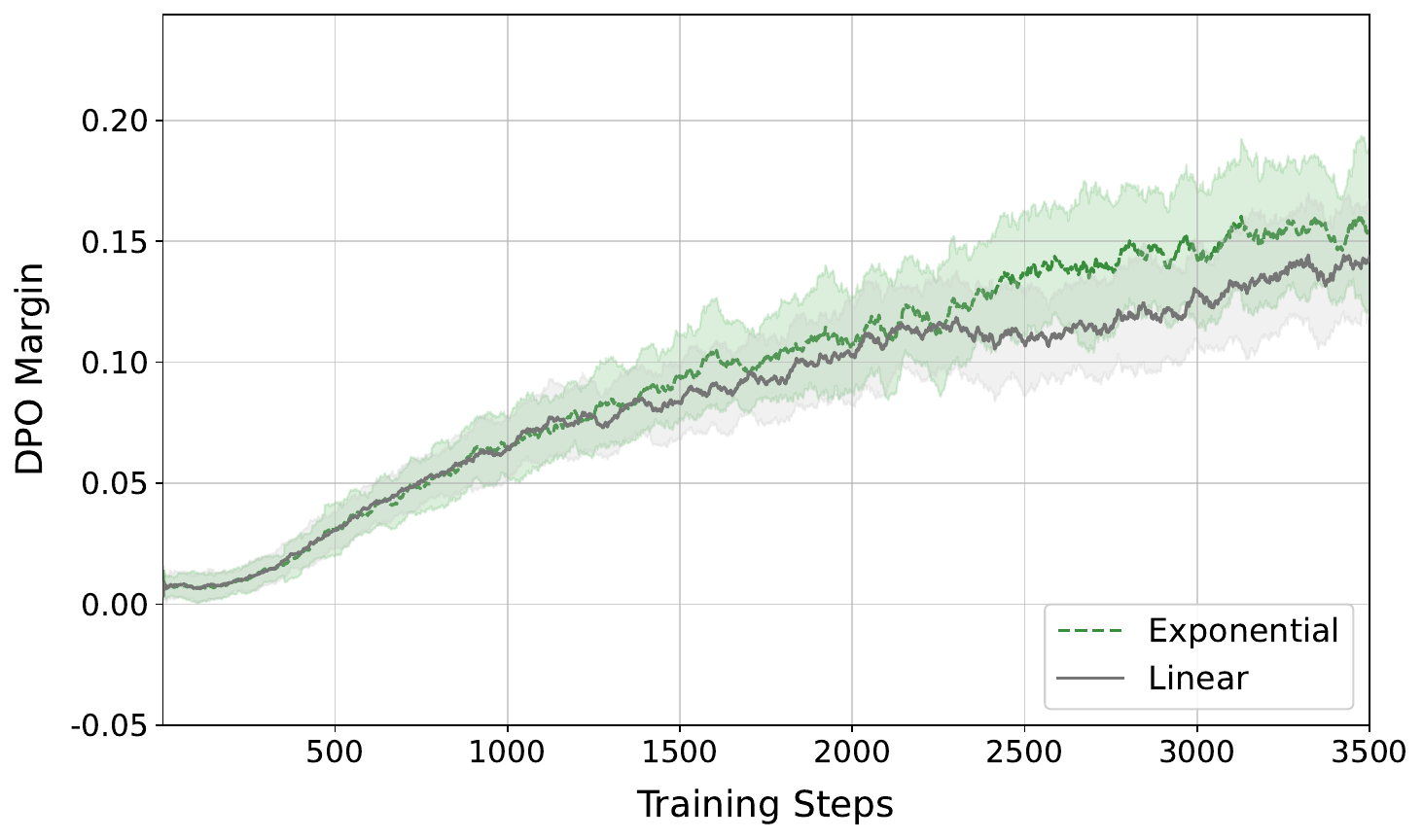}
    \caption{\textbf{Comparison of DPO Margin.}
    The exponential form (Green) maintains a higher margin than the linear form (Grey), indicating superior stability and learning discrimination.
    }
    \Description{}
    \label{fig:semantic_dpo_margin}
\end{figure}

We tracked the training dynamics of both variants under identical hyperparameters.
The comparison includes preference accuracy and DPO margin training dynamics.
As illustrated in Figure~\ref{fig:semantic_dpo_margin}, the exponential form presents a consistently higher DPO margin compared to the linear variant.
The DPO margin reflects the model confidence in distinguishing between preferred and rejected samples.
The lower margin of the linear variant suggests that soft suppression is insufficient to filter out the noise from conflicting pairs.
In contrast, the exponential form serves as a shaper filter, reducing the gradient contributions from pairs with semantic drift.
Thus, we choose the exponential form for our semantic-aware preference optimization.

Despite these advancements, our reliance on monocular depth and optical flow may still face challenges in scenes with severe occlusion or extremely rapid camera motion. 
Furthermore, our current physics parser categorizes events into coarse-grained quadrants. 


\begin{figure*}[t]
    \centering
    \begin{subfigure}[t]{0.33\textwidth}
        \centering
        \includegraphics[width=\textwidth]{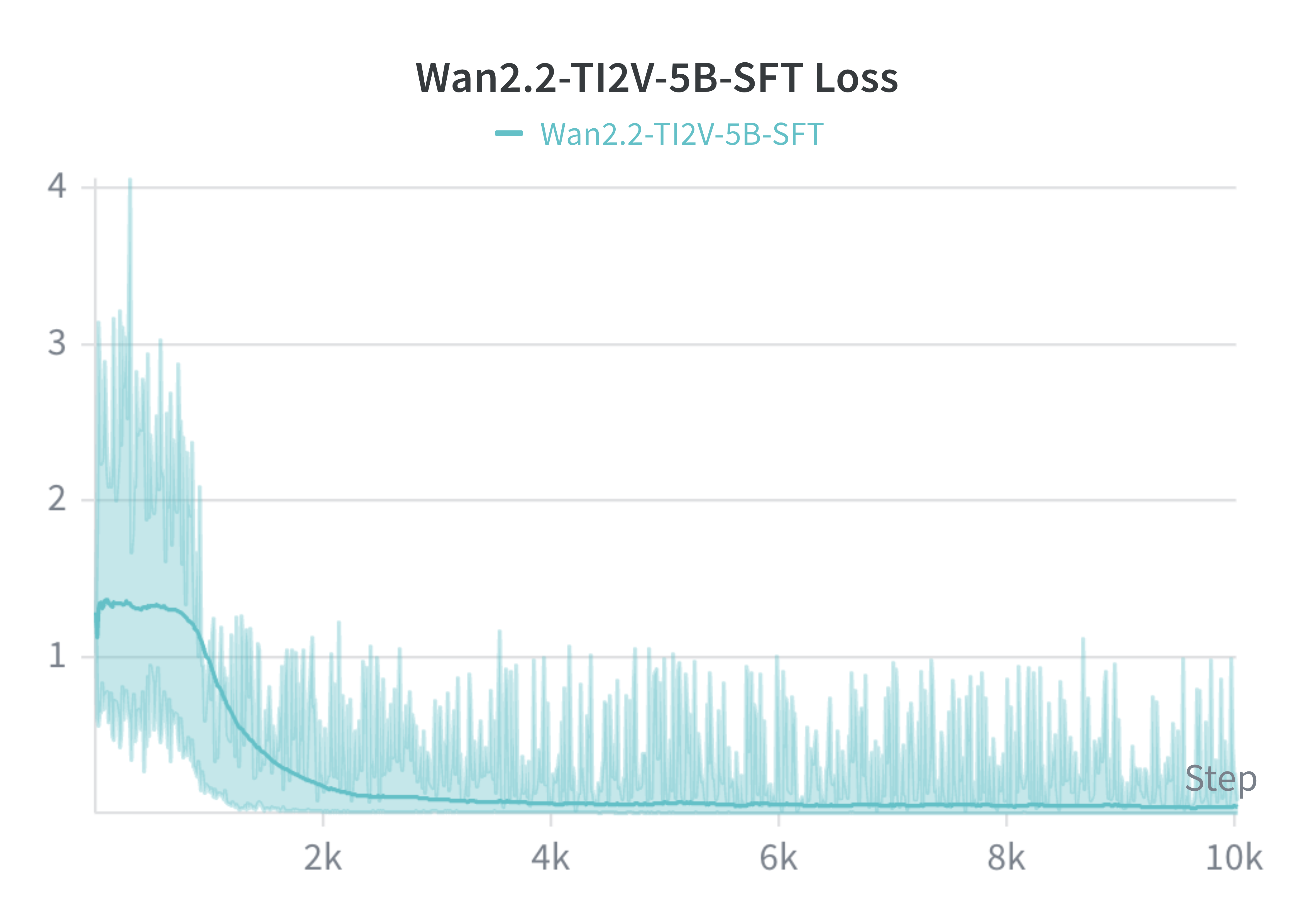}
        \caption{Wan2.2-TI2V-5B SFT loss.}
    \end{subfigure}
    \hfill
    \begin{subfigure}[t]{0.33\textwidth}
        \centering
        \includegraphics[width=\textwidth]{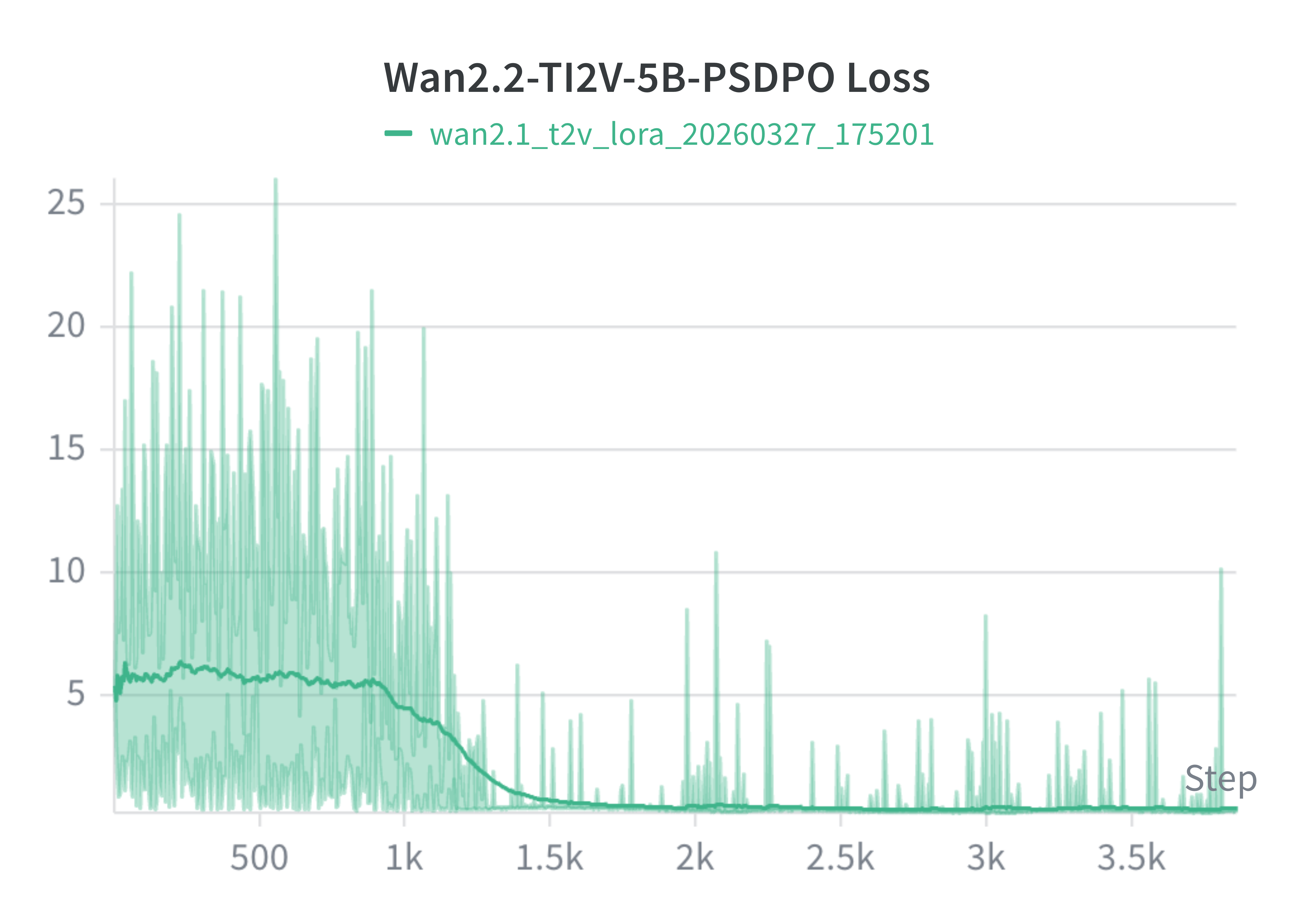}
        \caption{Wan2.2-TI2V-5B PSDPO loss.}
    \end{subfigure}
    \hfill
    \begin{subfigure}[t]{0.33\textwidth}
        \centering
        \includegraphics[width=\textwidth]{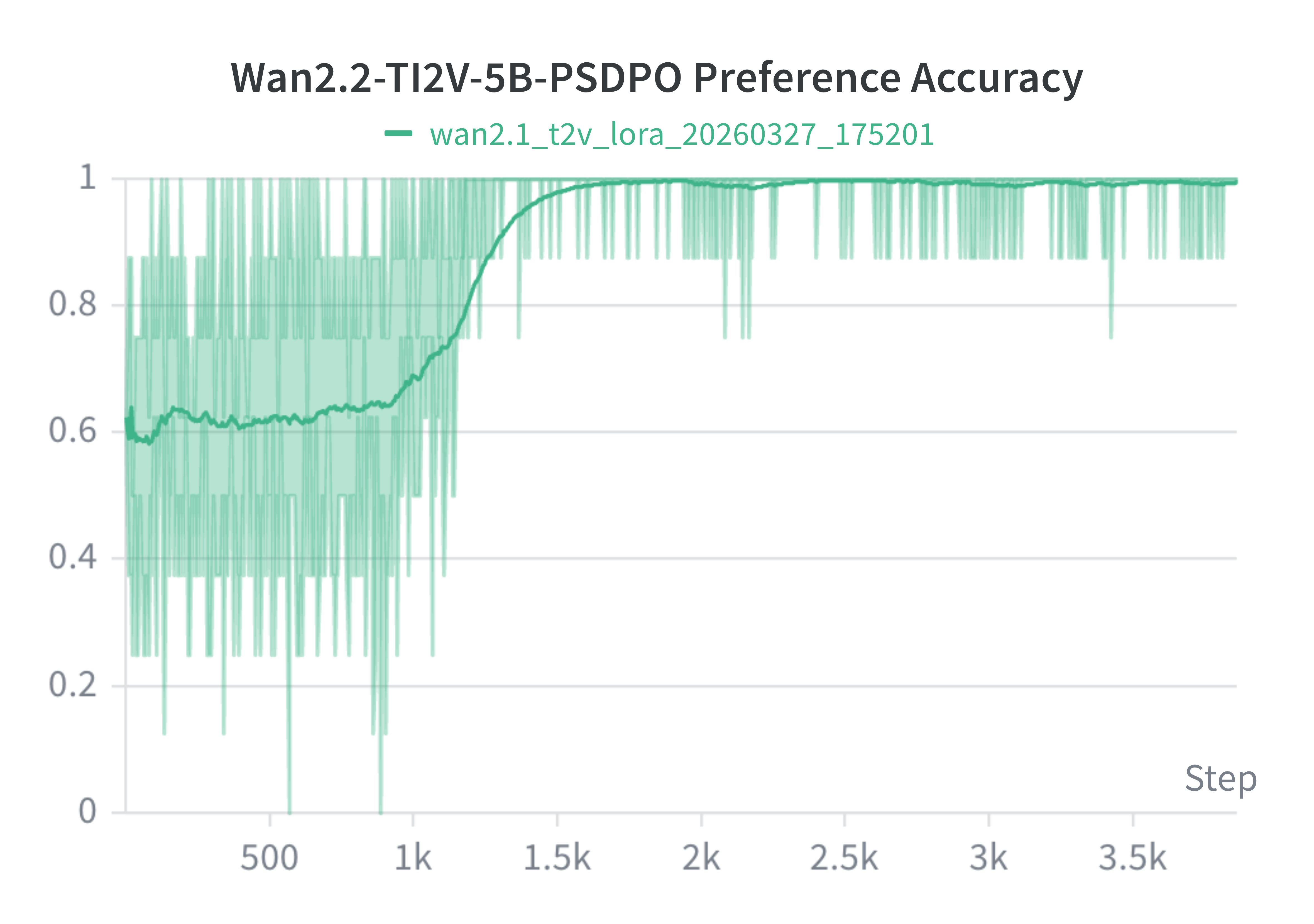}
        \caption{Wan2.2-TI2V-5B PSDPO preference accuracy.}
    \end{subfigure}
    \caption{
    Additional optimization curves on Wan2.2-TI2V-5B.
    (a) Supervised fine-tuning loss decreases smoothly and provides a stable initialization.
    (b) PSDPO loss drops rapidly after the early training stage and remains in a low regime.
    (c) Preference accuracy increases steadily and approaches saturation, indicating effective preference learning on an additional backbone.
    }
    \label{fig:wan22_extension}
\end{figure*}

\section{Extensibility of PSDPO on Wan2.2-5B model}
To further evaluate the extensibility of PSDPO beyond the Wan2.1-1.3B backbone, we additionally apply our training pipeline to \textbf{Wan2.2-TI2V-5B}. 
Our goal here is not to provide a full benchmark comparison across models, but to verify that PSDPO can be integrated into other text/image-to-video backbones with stable optimization behavior.

As shown in Figure~\ref{fig:wan22_extension}, the \textbf{SFT} stage on Wan2.2-TI2V-5B exhibits a smooth loss reduction and converges steadily, providing a stable initialization for subsequent preference optimization. 
After applying \textbf{PSDPO}, the training loss further decreases rapidly and reaches a low regime within a relatively small number of optimization steps. 
Meanwhile, the preference accuracy rises consistently and quickly approaches saturation, indicating that the model can effectively learn from the preference supervision under our physical-and-semantic weighting scheme.

These results suggest that PSDPO is not restricted to a specific architecture or model scale. 
Instead, it can serve as a general preference optimization strategy for video generation models, as long as the model supports standard supervised fine-tuning and pairwise preference learning. 
We emphasize that this experiment is intended to demonstrate \emph{training compatibility and optimization stability} on an additional backbone, which supports the practical extensibility of our method.

\section{Failure Mode Analysis}
\begin{figure*}[t]
    \centering
    \includegraphics[width=1\linewidth]{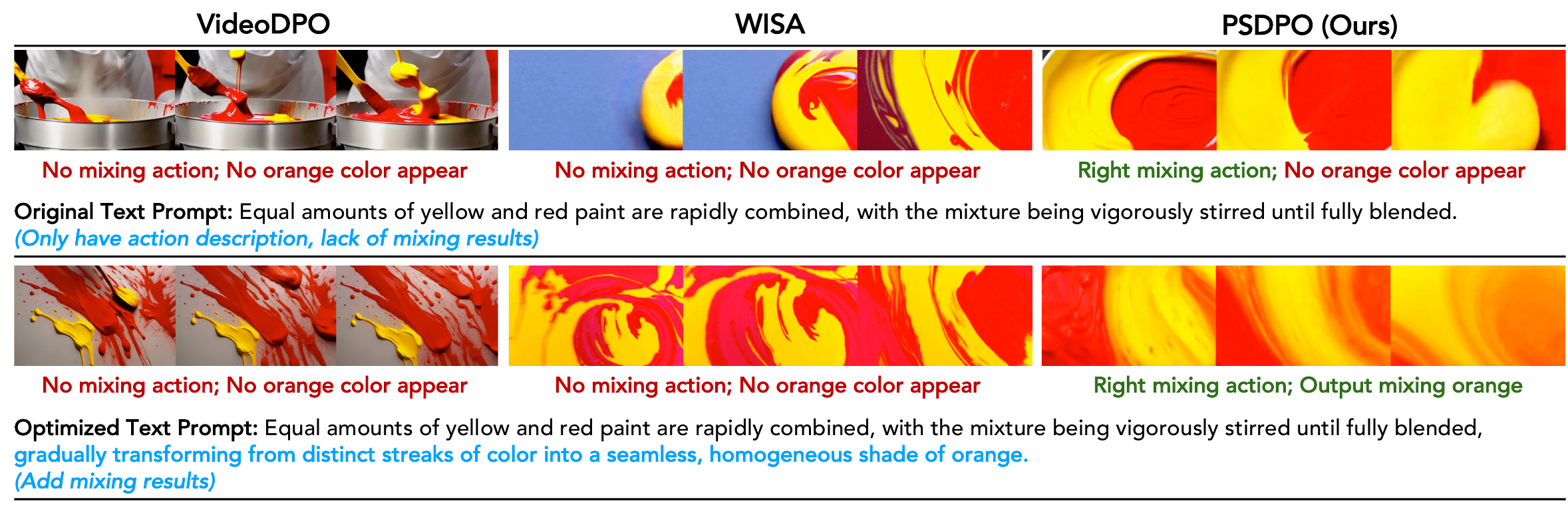}
    \caption{\textbf{Failure Case Analysis.}
    (Top) Under the implicit prompt, all models fail to infer the color change, revealing the dependency on data priors. 
    (Bottom) With an explicit prompt, PSDPO correctly generates the orange mixture, whereas baselines still suffer from semantic drift and fail to follow the state-change instruction.
    }
    \Description{}
    \label{fig:failure}
\end{figure*}

To better understand the boundaries of preference-based alignment, we analyze a representative failure case \textit{fluid color mixing}.
As shown in Figure~\ref{fig:failure}, with the original prompt, which describes the mixing action but implicitly omits the mixing result, all methods failed to generate the color transition.
This reveals a fundamental limitation that preference optimization aligns model capabilities with targets but does not inject new physical knowledge.
Since the underlying SFT model (Wan-1.3B) relies on data-driven correlation, it struggles to infer implicit physical consequence without explicit texture guidance. 
This confirms that physical plausibility in T2V generation remains heavily dependent on the comprehensiveness of the training data and the explicitness of the text prompt.

When the prompt is optimized to explicitly state the outcome, VideoDPO and WISA still fail to generate the correct orange color specified in the text instruction.
In contrast, PSDPO successfully generates both the vigorous mixing action and the correct color transition.
This validates the effectiveness of our semantic-aware modulation.
By dynamically penalizing semantic discrepancies during training, PSDPO ensures that the model remains sensitive to detailed texture instructions, ensuring correct physical behaviors.

\section{Limitations and Future works}
\label{appen:limitation}
While PSDPO achieves a state-of-the-art balance between physical plausibility and semantic fidelity, we acknowledge that PSDPO still faces challenges in unseen physical phenomena and reliance on data priors.

\noindent\textbf{Limitations.} 
As a preference optimization method, PSDPO cannot synthesize physical laws absent from the underlying distribution. 
Consequently, our performance is bounded by the capabilities of the base SFT model.
In this paper, we focus on 17 common physical phenomena, covering fundamental categories such as gravity, collision, combustion, and magnetism. 
We have not extensively accounted for rare or complex physical interactions outside this taxonomy. 
As observed in our failure case analysis \textit{color mixing}, when the base model lacks specific data priors for rare physics, preference optimization alone is insufficient to induce the correct dynamics without explicit guidance.

Additionally, our Physical Commonsense Evaluation (PCE) system relies on auxiliary geometric metrics, such as Depth and Optical Flow to ground MLLM reasoning. 
While robust for standard scenes, the reliability of these proxies is inextricably linked to the generation quality of the video model itself. 
In cases of severe visual artifacts or low-resolution generation, depth and flow estimations become noisy, potentially leading to inaccurate reward signals. 
Furthermore, current optical flow methods may degrade over long-video generation, where error accumulation in temporal consistency can limit the effectiveness of our metric-grounded evaluation for extended sequences.

\noindent\textbf{Future works.} 
Building on these insights, we have two promising directions for future research.
To mitigate the dependency on implicit SFT priors, we aim to explore methods that inject explicit physical constraints directly into the text prompts. By systematically augmenting prompts with descriptions of state changes or kinematic rules, we hope to guide the video generation model more effectively in scenarios where implicit commonsense is weak, bridging the gap between semantic instructions and physical execution.
We plan to extend current framework beyond the current closed-set of limited physical categories to open-world physical scenarios, potentially integrating LLM to access broader physical knowledge.

\clearpage
\bibliographystyle{plain}
\bibliography{reference}

@String{Computing = "Computing" }

@String{Chelsea = "Chelsea" }

@String(PAMI  = {IEEE Trans. Pattern Anal. Mach. Intell.})

@String(CVPR  = {IEEE Conf. Comput. Vis. Pattern Recog.})

@String(ICCV  = {Int. Conf. Comput. Vis.})

@String(ECCV  = {Eur. Conf. Comput. Vis.})

@String(NeurIPS = {Adv. Neural Inform. Process. Syst.})

@String(ICML  = {Int. Conf. Mach. Learn.})

@String(ICLR  = {Int. Conf. Learn. Represent.})

@String(IJCAI = {IJCAI})

@String(PAMI  = {IEEE TPAMI})

@String(CVPR  = {CVPR})

@String(ICCV  = {ICCV})

@String(ECCV  = {ECCV})

@String(NeurIPS = {NeurIPS})

@String(ICML  = {ICML})

@String(ICLR  = {ICLR})

@article{wan2025wan_arxiv2025,
  title={Wan: Open and advanced large-scale video generative models},
  author={Wan, Team and Wang, Ang and Ai, Baole and Wen, Bin and Mao, Chaojie and Xie, Chen-Wei and Chen, Di and Yu, Feiwu and Zhao, Haiming and Yang, Jianxiao and others},
  journal={arXiv preprint arXiv:2503.20314},
  year={2025}
}

@article{yang2024cogvideox_iclr2025,
  title={Cogvideox: Text-to-video diffusion models with an expert transformer},
  author={Yang, Zhuoyi and Teng, Jiayan and Zheng, Wendi and Ding, Ming and Huang, Shiyu and Xu, Jiazheng and Yang, Yuanming and Hong, Wenyi and Zhang, Xiaohan and Feng, Guanyu and others},
  journal={ICLR},
  year={2025}
}

@article{kong2024hunyuanvideo_arxiv2024,
  title={Hunyuanvideo: A systematic framework for large video generative models},
  author={Kong, Weijie and Tian, Qi and Zhang, Zijian and Min, Rox and Dai, Zuozhuo and Zhou, Jin and Xiong, Jiangfeng and Li, Xin and Wu, Bo and Zhang, Jianwei and others},
  journal={arXiv preprint arXiv:2412.03603},
  year={2024}
}

@article{rafailov2023direct_nips2023,
  title={Direct preference optimization: Your language model is secretly a reward model},
  author={Rafailov, Rafael and Sharma, Archit and Mitchell, Eric and Manning, Christopher D and Ermon, Stefano and Finn, Chelsea},
  journal={NeurIPS},
  year={2023}
}

@inproceedings{liu2025videodpo_cvpr2025,
  title={Videodpo: Omni-preference alignment for video diffusion generation},
  author={Liu, Runtao and Wu, Haoyu and Zheng, Ziqiang and Wei, Chen and He, Yingqing and Pi, Renjie and Chen, Qifeng},
  booktitle={CVPR},
  year={2025}
}

@article{bansal2024videophy_iclr2025,
  title={Videophy: Evaluating physical commonsense for video generation},
  author={Bansal, Hritik and Lin, Zongyu and Xie, Tianyi and Zong, Zeshun and Yarom, Michal and Bitton, Yonatan and Jiang, Chenfanfu and Sun, Yizhou and Chang, Kai-Wei and Grover, Aditya},
  journal={ICLR},
  year={2025}
}

@article{bansal2025videophy_arxiv2025,
  title={Videophy-2: A challenging action-centric physical commonsense evaluation in video generation},
  author={Bansal, Hritik and Peng, Clark and Bitton, Yonatan and Goldenberg, Roman and Grover, Aditya and Chang, Kai-Wei},
  journal={arXiv preprint arXiv:2503.06800},
  year={2025}
}

@inproceedings{huang2024vbench_cvpr2024,
  title={Vbench: Comprehensive benchmark suite for video generative models},
  author={Huang, Ziqi and He, Yinan and Yu, Jiashuo and Zhang, Fan and Si, Chenyang and Jiang, Yuming and Zhang, Yuanhan and Wu, Tianxing and Jin, Qingyang and Chanpaisit, Nattapol and others},
  booktitle={CVPR},
  year={2024}
}

@article{chen2025hierarchical_arxiv2025,
  title={Hierarchical fine-grained preference optimization for physically plausible video generation},
  author={Chen, Harold Haodong and Huang, Haojian and Chen, Qifeng and Yang, Harry and Lim, Ser-Nam},
  journal={NeurIPS},
  year={2025}
}

@article{zheng2023judging_nips2023,
  title={Judging llm-as-a-judge with mt-bench and chatbot arena},
  author={Zheng, Lianmin and Chiang, Wei-Lin and Sheng, Ying and Zhuang, Siyuan and Wu, Zhanghao and Zhuang, Yonghao and Lin, Zi and Li, Zhuohan and Li, Dacheng and Xing, Eric and others},
  journal={NeurIPS},
  year={2023}
}

@inproceedings{yang2024depth_cvpr2024,
  title={Depth anything: Unleashing the power of large-scale unlabeled data},
  author={Yang, Lihe and Kang, Bingyi and Huang, Zilong and Xu, Xiaogang and Feng, Jiashi and Zhao, Hengshuang},
  booktitle={CVPR},
  year={2024}
}

@inproceedings{xue2025phyt2v_cvpr2025,
  title={Phyt2v: Llm-guided iterative self-refinement for physics-grounded text-to-video generation},
  author={Xue, Qiyao and Yin, Xiangyu and Yang, Boyuan and Gao, Wei},
  booktitle={CVPR},
  year={2025}
}

@inproceedings{liu2024physgen_eccv2024,
  title={Physgen: Rigid-body physics-grounded image-to-video generation},
  author={Liu, Shaowei and Ren, Zhongzheng and Gupta, Saurabh and Wang, Shenlong},
  booktitle={ECCV},
  year={2024}
}

@article{tan2024physmotion_arxiv2024,
  title={Physmotion: Physics-grounded dynamics from a single image},
  author={Tan, Xiyang and Jiang, Ying and Li, Xuan and Zong, Zeshun and Xie, Tianyi and Yang, Yin and Jiang, Chenfanfu},
  journal={3DV},
  year={2026}
}

@article{wang2024internvid_iclr2024,
  title={Internvid: A large-scale video-text dataset for multimodal understanding and generation},
  author={Wang, Yi and He, Yinan and Li, Yizhuo and Li, Kunchang and Yu, Jiashuo and Ma, Xin and Li, Xinhao and Chen, Guo and Chen, Xinyuan and Wang, Yaohui and others},
  journal={ICLR},
  year={2024}
}

@article{wang2025wisa_arxiv2025,
  title={Wisa: World simulator assistant for physics-aware text-to-video generation},
  author={Wang, Jing and Ma, Ao and Cao, Ke and Zheng, Jun and Zhang, Zhanjie and Feng, Jiasong and Liu, Shanyuan and Ma, Yuhang and Cheng, Bo and Leng, Dawei and others},
  journal={NeurIPS},
  year={2025}
}

@misc{openai2024sora,
  title        = {Video Generation Models as World Simulators},
  author       = {{OpenAI}},
  year         = {2024},
  howpublished = {\url{https://openai.com/research/video-generation-models-as-world-simulators}},
  note         = {Accessed 2024}
}

@article{unterthiner2019fvd,
  title={FVD: A new metric for video generation},
  author={Unterthiner, Thomas and Van Steenkiste, Sjoerd and Kurach, Karol and Marinier, Rapha{\"e}l and Michalski, Marcin and Gelly, Sylvain},
  journal={ICLRW},
  year={2019}
}

@article{xu2021videoclip_emnlp2021,
  title = "{V}ideo{CLIP}: Contrastive Pre-training for Zero-shot Video-Text Understanding",
  author={Xu, Hu and Ghosh, Gargi and Huang, Po-Yao and Okhonko, Dmytro and Aghajanyan, Armen and Metze, Florian and Zettlemoyer, Luke and Feichtenhofer, Christoph},
  journal={EMNLP},
  year={2021}
}

@article{wang2025physcorr_arxiv2025,
  title={PhysCorr: Dual-Reward DPO for Physics-Constrained Text-to-Video Generation with Automated Preference Selection},
  author={Wang, Peiyao and Wang, Weining and Li, Qi},
  journal={arXiv preprint arXiv:2511.03997},
  year={2025}
}

@article{wu2025densedpo_arxiv2025,
  title={DenseDPO: Fine-Grained Temporal Preference Optimization for Video Diffusion Models},
  author={Wu, Ziyi and Kag, Anil and Skorokhodov, Ivan and Menapace, Willi and Mirzaei, Ashkan and Gilitschenski, Igor and Tulyakov, Sergey and Siarohin, Aliaksandr},
  journal={NeurIPS},
  year={2025}
}

@article{jiang2025huvidpo_arxiv2025,
  title={Huvidpo: Enhancing video generation through direct preference optimization for human-centric alignment},
  author={Jiang, Lifan and Wu, Boxi and Zhang, Jiahui and Guan, Xiaotong and Chen, Shuang},
  journal={arXiv preprint arXiv:2502.01690},
  year={2025}
}

@article{huang2025vistadpo_icml2025,
  title={VistaDPO: Video Hierarchical Spatial-Temporal Direct Preference Optimization for Large Video Models},
  author={Huang, Haojian and Chen, Haodong and Wu, Shengqiong and Luo, Meng and Fu, Jinlan and Du, Xinya and Zhang, Hanwang and Fei, Hao},
  journal={ICML},
  year={2025}
}

@article{barratt2018note_arxiv2018,
  title={A note on the inception score},
  author={Barratt, Shane and Sharma, Rishi},
  journal={arXiv preprint arXiv:1801.01973},
  year={2018}
}

@inproceedings{teed2020raft_eccv2020,
  title={Raft: Recurrent all-pairs field transforms for optical flow},
  author={Teed, Zachary and Deng, Jia},
  booktitle={ECCV},
  year={2020}
}

@article{hu2022lora_iclr2022,
  title={Lora: Low-rank adaptation of large language models.},
  author={Hu, Edward J and Shen, Yelong and Wallis, Phillip and Allen-Zhu, Zeyuan and Li, Yuanzhi and Wang, Shean and Wang, Lu and Chen, Weizhu and others},
  journal={ICLR},
  year={2022}
}

@article{loshchilov2017decoupled_arxiv2017,
  title={Decoupled weight decay regularization},
  author={Loshchilov, Ilya and Hutter, Frank},
  journal={ICLR},
  year={2019}
}

@article{zhang2025think_arxiv2025,
  title={Think Before You Diffuse: LLMs-Guided Physics-Aware Video Generation},
  author={Zhang, Ke and Xiao, Cihan and Mei, Yiqun and Xu, Jiacong and Patel, Vishal M},
  journal={arXiv preprint arXiv:2505.21653},
  year={2025}
}

@article{qian2025rdpo,
  title={RDPO: Real Data Preference Optimization for Physics Consistency Video Generation},
  author={Qian, Wenxu and Wang, Chaoyue and Peng, Hou and Tan, Zhiyu and Li, Hao and Zeng, Anxiang},
  journal={arXiv preprint arXiv:2506.18655},
  year={2025}
}

@article{christiano2017deep_nips2017,
  title={Deep reinforcement learning from human preferences},
  author={Christiano, Paul F and Leike, Jan and Brown, Tom and Martic, Miljan and Legg, Shane and Amodei, Dario},
  journal={NeurIPS},
  year={2017}
}

@article{meng2024towards_icml2025,
  title={Towards world simulator: Crafting physical commonsense-based benchmark for video generation},
  author={Meng, Fanqing and Liao, Jiaqi and Tan, Xinyu and Shao, Wenqi and Lu, Quanfeng and Zhang, Kaipeng and Cheng, Yu and Li, Dianqi and Qiao, Yu and Luo, Ping},
  journal={ICML},
  year={2025}
}

@article{kang2024far_icml2025,
  title={How far is video generation from world model: A physical law perspective},
  author={Kang, Bingyi and Yue, Yang and Lu, Rui and Lin, Zhijie and Zhao, Yang and Wang, Kaixin and Huang, Gao and Feng, Jiashi},
  journal={ICML},
  year={2025}
}

@inproceedings{xie2024physgaussian_cvpr2024,
  title={Physgaussian: Physics-integrated 3d gaussians for generative dynamics},
  author={Xie, Tianyi and Zong, Zeshun and Qiu, Yuxing and Li, Xuan and Feng, Yutao and Yang, Yin and Jiang, Chenfanfu},
  booktitle={CVPR},
  year={2024}
}

@article{Zhang2026PhysRVGPU_arxiv2026,
  title={PhysRVG: Physics-Aware Unified Reinforcement Learning for Video Generative Models},
  author={Qiyuan Zhang and Biao Gong and Shuai Tan and Zheng Zhang and Yujun Shen and Xing Zhu and Yuyuan Li and Kelu Yao and Chunhua Shen and Changqing Zou},
  journal={arXiv preprint arXiv:2601.11087},
  year={2026}
}

@article{hacohen2024ltx_arxiv2024,
  title={Ltx-video: Realtime video latent diffusion},
  author={HaCohen, Yoav and Chiprut, Nisan and Brazowski, Benny and Shalem, Daniel and Moshe, Dudu and Richardson, Eitan and Levin, Eran and Shiran, Guy and Zabari, Nir and Gordon, Ori and others},
  journal={arXiv preprint arXiv:2501.00103},
  year={2024}
}

@article{hao2025enhancing_arxiv2025,
  title={Enhancing physical plausibility in video generation by reasoning the implausibility},
  author={Hao, Yutong and Chen, Chen and Mian, Ajmal Saeed and Xu, Chang and Liu, Daochang},
  journal={arXiv preprint arXiv:2509.24702},
  year={2025}
}

@article{le2025gravity_arxiv2025,
  title={What about gravity in video generation? Post-Training Newton's Laws with Verifiable Rewards},
  author={Le, Minh-Quan and Zhu, Yuanzhi and Kalogeiton, Vicky and Samaras, Dimitris},
  journal={arXiv preprint arXiv:2512.00425},
  year={2025}
}

@article{yuan2026inference_arxiv2026,
  title={Inference-time Physics Alignment of Video Generative Models with Latent World Models},
  author={Yuan, Jianhao and Zhang, Xiaofeng and Friedrich, Felix and Beltran-Velez, Nicolas and Hall, Melissa and Askari-Hemmat, Reyhane and Han, Xiaochuang and Ballas, Nicolas and Drozdzal, Michal and Romero-Soriano, Adriana},
  journal={arXiv preprint arXiv:2601.10553},
  year={2026}
}

@article{ji2025physmaster_arxiv2025,
  title={Physmaster: Mastering physical representation for video generation via reinforcement learning},
  author={Ji, Sihui and Chen, Xi and Tao, Xin and Wan, Pengfei and Zhao, Hengshuang},
  journal={arXiv preprint arXiv:2510.13809},
  year={2025}
}

@article{ding2025understanding,
  title={Understanding world or predicting future? a comprehensive survey of world models},
  author={Ding, Jingtao and Zhang, Yunke and Shang, Yu and Zhang, Yuheng and Zong, Zefang and Feng, Jie and Yuan, Yuan and Su, Hongyuan and Li, Nian and Sukiennik, Nicholas and others},
  journal={ACM Computing Surveys},
  year={2025}
}

@article{yang2025qwen3,
  title={Qwen3 technical report},
  author={Yang, An and Li, Anfeng and Yang, Baosong and Zhang, Beichen and Hui, Binyuan and Zheng, Bo and Yu, Bowen and Gao, Chang and Huang, Chengen and Lv, Chenxu and others},
  journal={arXiv preprint arXiv:2505.09388},
  year={2025}
}

@article{mai2025contextanyone,
  title={ContextAnyone: Context-Aware Diffusion for Character-Consistent Text-to-Video Generation},
  author={Mai, Ziyang and Tai, Yu-Wing},
  journal={arXiv preprint arXiv:2512.07328},
  year={2025}
}

@inproceedings{ke2021musiq_iccv2021,
  title={Musiq: Multi-scale image quality transformer},
  author={Ke, Junjie and Wang, Qifei and Wang, Yilin and Milanfar, Peyman and Yang, Feng},
  booktitle={ICCV},
  year={2021}
}

@inproceedings{radford2021learning_icml2021,
  title={Learning transferable visual models from natural language supervision},
  author={Radford, Alec and Kim, Jong Wook and Hallacy, Chris and Ramesh, Aditya and Goh, Gabriel and Agarwal, Sandhini and Sastry, Girish and Askell, Amanda and Mishkin, Pamela and Clark, Jack and others},
  booktitle={ICML},
  year={2021}
}

@inproceedings{hessel2021clipscore_emnlp2021,
  title={Clipscore: A reference-free evaluation metric for image captioning},
  author={Hessel, Jack and Holtzman, Ari and Forbes, Maxwell and Le Bras, Ronan and Choi, Yejin},
  booktitle={EMNLP},
  year={2021}
}

@inproceedings{liu2024evalcrafter_cvpr2024,
  title={Evalcrafter: Benchmarking and evaluating large video generation models},
  author={Liu, Yaofang and Cun, Xiaodong and Liu, Xuebo and Wang, Xintao and Zhang, Yong and Chen, Haoxin and Liu, Yang and Zeng, Tieyong and Chan, Raymond and Shan, Ying},
  booktitle={CVPR},
  year={2024}
}

@article{meng2025grounding_ijcai2025,
  title={Grounding creativity in physics: A brief survey of physical priors in aigc},
  author={Meng, Siwei and Luo, Yawei and Liu, Ping},
  journal={IJCAI},
  year={2025}
}

@inproceedings{ren2018learning_ciml2018,
  title={Learning to reweight examples for robust deep learning},
  author={Ren, Mengye and Zeng, Wenyuan and Yang, Bin and Urtasun, Raquel},
  booktitle={ICML},
  year={2018},
}

@inproceedings{han2018co_nips2018,
  title={Co-teaching: Robust training of deep neural networks with extremely noisy labels},
  author={Han, Bo and Yao, Quanming and Yu, Xingrui and Niu, Gang and Xu, Miao and Hu, Weihua and Tsang, Ivor and Sugiyama, Masashi},
  booktitle={NeurIPS},
  year={2018}
}

@inproceedings{bengio2009curriculum_icml2009,
  title={Curriculum learning},
  author={Bengio, Yoshua and Louradour, J{\'e}r{\^o}me and Collobert, Ronan and Weston, Jason},
  booktitle={ICML},
  year={2009}
}

@ARTICLE{9392296_pami2022,
  author={Wang, Xin and Chen, Yudong and Zhu, Wenwu},
  journal={IEEE Transactions on Pattern Analysis and Machine Intelligence}, 
  title={A Survey on Curriculum Learning}, 
  booktitle={PAMI},
  year={2022},
  }

@article{meng2025phymagic_eccv2026,
  title={PhyMAGIC: Physical motion-aware generative inference with confidence-guided LLM},
  author={Meng, Siwei and Luo, Yawei and Liu, Ping},
  journal={ECCV},
  year={2026}
}

\end{document}